\documentclass[10pt,twocolumn,letterpaper]{article}

\usepackage{cvpr}
\usepackage{amsmath,amssymb,amsfonts}
\usepackage{subfigure}
\usepackage{graphicx}
\usepackage{textcomp}
\usepackage{algorithm}
\usepackage{algorithmic}
\usepackage{microtype}
\usepackage{booktabs}
\usepackage{color}
\newcommand{\D}[2]{\frac{\partial #1}{\partial #2}}

\newcommand{\hl}[1]{\textcolor{black}{#1}}

\usepackage[pagebackref=true,breaklinks=true,letterpaper=true,colorlinks,bookmarks=false]{hyperref}

\cvprfinalcopy 

\def\cvprPaperID{5394} 

\ifcvprfinal\fi
\begin{document}

\title{Learning Multi-Layered GBDT Via Back Propagation}

\author{Zhendong Zhang\\
School of Electronic Engineering, Xidian University\\
Xi'an, 710071, China\\
{\tt\small zhd.zhang.ai@gmail.com}
}

\maketitle

\begin{abstract}
	Deep neural networks are able to learn multi-layered representation via back propagation (BP). Although the gradient boosting decision tree (GBDT) is effective for modeling tabular data, it is non-differentiable with respect to its input, thus suffering from learning multi-layered representation. In this paper, we propose a framework of learning multi-layered GBDT via BP. We approximate the gradient of GBDT based on linear regression. Specifically, we use linear regression to replace the constant value at each leaf ignoring the contribution of individual sample to the tree structure. In this way, we estimate the gradient for intermediate representations, which facilitates BP for multi-layered GBDT. Experiments show the effectiveness of the proposed method in terms of performance and representation ability. To the best of our knowledge, this is the first work of optimizing multi-layered GBDT via BP. \hl{This work provides a new possibility of exploring deep tree based learning and combining GBDT with neural networks.} 
\end{abstract}

\section{Introduction}
\label{introduction}
Deep neural networks have achieved outstanding breakthrough of machine learning in recent years \cite{goodfellow2016deep}. Feature engineering is important for traditional machine learning methods that lack the ability to learn intermediate representation from raw data. However, deep neural networks can learn good representations from raw data \cite{bengio2013representation}, which is believed to be one of the key advantages beyond traditional methods. This is achieved by constructing multi-layered neural networks and learning hierarchical representations. Although many learning strategies have been proposed for training deep neural networks such as target propagation \cite{lee2015difference} and layer-wise greedy training \cite{hinton2012practical}, back propagation (BP) \cite{rumelhart1986learning} with stochastic gradient descent is still the dominant approach. BP is used to compute the gradients of parameters layer by layer in the inverse direction. On the other hand, tree based ensembles such as random forest \cite{breiman2001random} and gradient boosting decision trees (GBDT) \cite{friedman2001greedy, chen2016xgboost} \hl{(we just name a few representative works)} are still the best choice of modeling tabular data because of not only their performance but also their computational efficiency and interpretability. However, there are no hierarchical representations of these models which limit their expressiveness. How can we construct multi-layered tree ensembles?

Recently, researchers have explored this issue popularly. Zhou and Feng \cite{zhou2017deep} proposed a deep forest framework, which was the first attempt to construct a multi-layered model using tree ensembles. They introduced fine-grained scanning and cascading operations and constructed a multi-layered structure with adaptive model complexity. However, the representation learning ability of deep forest was difficult to explicitly examine. Later, Feng et al. \cite{feng2018multi} proposed the first multi-layered structure called mGBDT using gradient boosting decision trees as building blocks per layer with an explicit emphasis on its representation learning ability. mGBDT was learned via target propagation. Specifically, targets were approximated and propagated by a set of inverse functions corresponding to the forward functions. Both inverse and forward functions were constructed by GBDT. Such explorations showed the feasibility of learning hierarchical tree ensembles. Their results demonstrated that tree ensembles were beneficial from multi-layered representations. Several methods of differentiable and probabilistic decision trees have been proposed \cite{frosst2017distilling, kontschieder2015deep, Popov2020Neural, hazimeh2020tree, tanno2019adaptive,feng2020soft}. The probability of going to the left branch was determined by the inner product between inputs and learned tree parameters. Training this kind of decision trees was similar to training neural networks. For traditional decision trees used in GBDT, the left branch and right branch were split by a threshold along one of the features in a deterministic manner. Since probability decision trees are different from the ones used in GBDT in terms of hypothesis space and training methodology, \cite{frosst2017distilling, kontschieder2015deep, Popov2020Neural, hazimeh2020tree,tanno2019adaptive,feng2020soft} are outside the scope of this work. 

\hl{
	Researchers also have explored combining GBDT with neural networks. \cite{ke2019deepgbm} distills the learned representations of GBDT into deep neural networks. In \cite{ivanov2020boost}, the inputs are first processed by GBDT then processed by graph neural networks. GBDT is learned via the gradients provided by graph neural networks. Although the accuracy is improved, the combination mechanism is restricted.
}

In this work, we focus on traditional decision trees. We learn multi-layered GBDT via BP which is different from mGBDT. Since GBDT is non-differentiable and non-parametric, it seems impossible to apply BP to GBDT. Our solution to this problem is as follows: we replace the constant value at each leaf with linear regression, called piece-wise linear GBDT (GBDT-PL) in a recent work \cite{shi2018gradient}. We first learn the structure of each regression tree, and use it to decide the set of inputs belonging to each leaf. We apply linear regression to each leaf to obtain the optimal linear parameters $\mathbf{w}$. Then, we express the gradient of output w.r.t its input in terms of $\mathbf{w}$ (see Section \ref{sec:method} for details). Note that this is an approximated gradient because the true gradient depends on not only the linear parameters but also the structure of trees. When samples for training become massive, it is reasonable to assume individual sample has little effect on the structure of trees. In fact, a series of works indicates that the true gradient is not necessary for updating parameters \cite{bengio2013estimating, czarnecki2017understanding, maddison2016concrete}. When we stack multi layers of GBDT, the gradient information can be propagated from the final output to the input layer by layer via BP. Then, we update the variables of hidden layers along the direction of gradient descent and re-train GBDT to fit those updated hidden variables. Although we could compute the gradient of $\mathbf{w}$ in a similar way, it is unnecessary to do that during the BP process. We obtain $\mathbf{w}$ by linear regression after the structure of trees is learned, instead of gradient descent. We call the proposed method as GBDT-BP. We evaluate GBDT-BP in terms of performance and representation learning ability on both synthetic data and real-world data. Compared with existing methods, our main contributions and novelties are as follows:
\begin{itemize}
	\item Although the idea of combining GBDT and linear regression at leaves appears earlier, this is the first work to explicitly point out that such a combination leads to an approximately differentiable model.
	\item This is the first work to learn multi-layered GBDT via BP which provides a new possibility of exploring deep tree based learning and \hl{combining GBDT with neural networks. In principle, GBDT-BP and neural networks can be combined in arbitrary way}.
\end{itemize}

\section{Background}

\subsection{Back Propagation}
Back propagation (BP), shorthand for "backward propagation of errors", is a method to compute the gradients w.r.t. the weights of multi-layered neural networks which has become the key ingredient of the training algorithm of deep neural networks. Although BP is originally proposed for neural networks, in principle it can be utilized to compute the gradients of any layer structured differentiable models. Thus, we formalize the process of BP beyond neural networks as follows.

Denote $\mathcal{L}$ as the final loss and $\mathbf{h}_i$ as the hidden vector of $i$-th layer. Denote $f_{\mathbf{w}_i}$ as the feed forward function of $i$-th layer parameterized by $\mathbf{w}_i$. That is, $\mathbf{h}_i = f_{\mathbf{w}_{i-1}}(\mathbf{h}_{i-1})$. Given the gradients of $\mathbf{h}_i$, BP computes the gradients of $\mathbf{h}_{i-1}$ and $\mathbf{w}_{i-1}$ as follows:
\begin{equation}
\label{eq_bp_h}
\D{\mathcal{L}}{\mathbf{h}_{i-1}} = \left(\D{\mathbf{h}_i}{\mathbf{h}_{i-1}} \right)^T \D{\mathcal{L}}{\mathbf{h}_i}
\end{equation}
\begin{equation}
\label{eq_bp_w}
\D{\mathcal{L}}{\mathbf{w}_{i-1}} = \left(\D{\mathbf{h}_i}{\mathbf{w}_{i-1}} \right)^T \D{\mathcal{L}}{\mathbf{h}_i}
\end{equation}
where $\D{\mathbf{h}_i}{\mathbf{h}_{i-1}}$ and $\D{\mathbf{h}_i}{\mathbf{w}_{i-1}}$ are jacobians. From \eqref{eq_bp_h}, the gradients of the $i$-th layer are propagated to the $(i-1)$-th layer. Then, the same rule is applied layer by layer until the inputs. In this way, the gradients of $\mathbf{w}$ at all layers are obtained.

BP requires that the model is differentiable or the gradient can be evaluated within acceptable computational complexity. A series of works is proposed to overcome these limitations of BP in particular situations. \cite{bengio2013estimating} proposed straight through estimator to learn binary variables. \cite{maddison2016concrete} proposed relaxation methods to learn a particular discrete probability distribution. \cite{czarnecki2017understanding} proposed the synthetic gradient which reduces the computation and communication complexity for distributed training. Their model is updated using the synthetic gradient instead of the true gradient. The effectiveness of these works indicates that it is a good choice to use the approximated gradient when the model is non-differentiable or the true gradient is too expensive to evaluate.

\subsection{Gradient Boosting Decision Trees}
Boosting is a widely used strategy for ensemble learning \cite{zhou2015ensemble}. Boosting algorithms iteratively add weak learners into the final strong learner. The construction of each weak learner is usually guided by the gradient of current predictions, i.e. gradient boosting. When we choose decision trees as the weak learner, we get gradient boosting decision trees. Formally, denote $\mathcal{M}$ as the function of an individual tree. The prediction of GBDT is the summation of all trees:
\begin{equation}
\hat{y} = \sum_{i=1}^{N} \mathcal{M}_i(x)
\end{equation}
Given the first $N$ trees, the $(N+1)$-th tree is grown by fitting current negative gradient w.r.t. $\hat{y}$. GBDT is first proposed in \cite{friedman2001greedy}. It is still the best choice for modeling tabular data which spans a broad range of real-world applications.

The most important problem of GBDT is how to split a node. The quality of splitting is measured by some objectives. Thus, this problem is cast to choosing suitable objectives. Traditional GBDTs use so-called ``friedman mse''. Recent implementations for GBDTs such as XGBoost \cite{chen2016xgboost} and LightGBM \cite{ke2017lightgbm} use not only the first order gradient but also the second order gradient to guide the learning of each tree. \cite{chen2016xgboost} firstly derived the corresponding splitting objective which is a generalization of ``friedman mse''. When the second order gradient is a constant, their proposed objective is equivalent of ``friedman mse''.

Another problem of GBDT is how to compute the predictions of each leaf. A constant value is usually used. For traditional GBDT, this value equals the mean of the negative gradient as follows:
\begin{equation}
\label{eq:leaf1}
\frac{1}{N} \sum_{i\in leaf} - g_i
\end{equation}
where $N$ is the number of samples belonging to the same leaf and  $g$ is the gradient. For implementations using second order gradient, the leaf value equals to the negative gradient weighted by second order gradient.
\begin{equation}
\label{eq:leaf2}
\frac{-\sum_{i\in leaf}g_i}{\lambda + \sum_{i\in leaf} g'_i}
\end{equation}
where $g'$ is the second order gradient and $\lambda$ is the strength of $L_2$ regularization. \cite{de2017linxgboost} replaced the constant leaf value with linear regression to derive the corresponding spitting objective.

\begin{figure*}[t]
	\centering
	\includegraphics[width=0.7\textwidth]{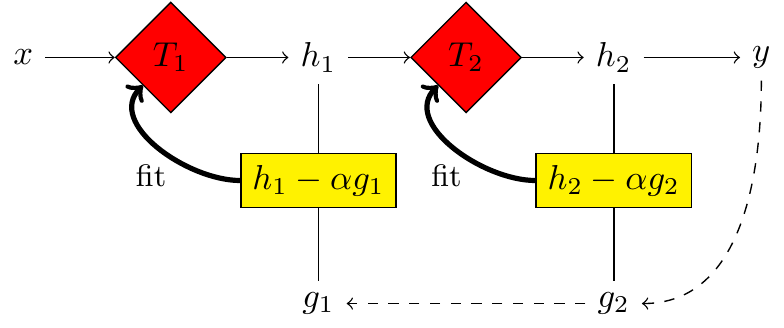}
	\caption{Flow of GBDT-BP with two hidden layers. We update the hidden variables via gradient descant and retrain decision trees to fit the updated ones. $x$ and $y$ are the input and output respectively. $h$ is the hidden representation and $g$ is its gradient. $\alpha$ is the learning rate, while $T$ is the GBDT model. The subscripts 1 and 2 indicate the number of layers.}
	\label{fig:flow}
\end{figure*}

\section{Proposed Method}
\label{sec:method}
In this section, we describe the proposed method in a top-down manner. We first provide the overall learning algorithm. Then, we show how to compute the gradient of GBDT. Finally, we describe how to learn the linear regression at each node. Implementation details are also discussed.

\subsection{Overall Learning Algorithm}
\label{sec:overall}
Consider a multi-layered structure with $L-1$ hidden layers. Denote $\mathbf{h}_i \in \mathbf{R}^{d_i}$ where $i\in \{0, 1, 2, \dots L\}$ as the hidden variables of $i$-th layer with dimension $d_i$. For convenience, denote $\mathbf{h}_0$ and $\mathbf{h}_L$ as the input and the final output layer respectively. Except for the input layer, each element of $\mathbf{h}_i$ is obtained by a GBDT with $\mathbf{h}_{i-1}$ as its input. That is, $\mathbf{h}_i$ is a concatenation of the outputs of $d_i$ GBDTs. Note that we replace the constant prediction at each leaf with linear regression to make GBDT approximate differentiable.

Given a differentiable loss $\mathcal{L}$, the gradient of $L$-th layer is obtained directly. It is required to compute the gradient of all hidden layers. Suppose that the jacobian of any two adjacent layers is given (we will describe how to compute it for GBDT in Section \ref{sec:gradient}). Then, we can compute the gradient of all hidden layers via BP as in~\eqref{eq_bp_h}. Although the BP process is similar to deep neural networks, the update rule is significantly different. For neural networks, their parameters are updated directly by (stochastic) gradient descent. Recall that there are linear parameters learned at each leaf and thus the model becomes parametric. However, it is meaningless to apply gradient descent to these parameters. It is meaningful only when the tree structure is fixed. Unfortunately, the tree structure is dynamically changed during training. Thus, it is unnecessary to compute the gradient of the linear parameters and update them. We only update the hidden variables via gradient descent as follows:
\begin{equation}
\label{eq:update}
\mathbf{h}_i \leftarrow \mathbf{h}_i - \alpha\D{\mathcal{L}}{\mathbf{h}_i}
\end{equation}
where $\alpha$ is the learning rate. Then, we clean previously learned GBDTs and train new ones from scratch to fit the updated hidden variables based on mean square error (MSE). Note that when we re-train new GBDTs to fit the updated $\mathbf{h}_i$, $\mathbf{h}_{i-1}$ has not been updated yet. That is, hidden variables and GBDTs are updated alternately layer by layer in the inverse direction. We summarize GBDT-BP in Algorithm~\ref{arg:learn} and provide the flow of GBDT-BP with two hidden layers in Fig.~\ref{fig:flow}.

\begin{algorithm*}[t]
	\caption{Learning multi-layered GBDT via back propagation}
	\label{arg:learn}
	\begin{algorithmic}
		\STATE {\bfseries Input:} Number of layers $L$, training dataset $\{\mathbf{h}_0, \mathbf{Y}\}$, final loss function $\mathcal{L}$, learning rate $\alpha$ for hidden variables, learning rate $\beta$ for GBDT and number of boosters $N_i$
		\STATE {\bfseries Output:} Multi-layered GBDTs
		\STATE Initialize forward function $\mathcal{T}_i$ of layer $i$ for $i=0, 1,\dots L-1$
		
		\REPEAT
		\STATE // Forward for computing hidden variables
		
		\FOR{$i=0$ {\bfseries to} $L-1$}
		\STATE $\mathbf{h}_{i+1} = \mathcal{T}_i(\mathbf{h}_i)$
		\ENDFOR
		\STATE // Backward for computing gradients
		\STATE $\mathbf{g}_L \leftarrow \D{\mathcal{L}(\mathbf{h}_L, \mathbf{Y})}{\mathbf{h}_L}$
		
		\FOR{$i=L$ {\bfseries to} $1$}
		\STATE $\mathbf{g}_{i-1} \leftarrow \left(\D{\mathbf{h}_i}{\mathbf{h}_{i-1}}\right)^T\mathbf{g}_i$ // How to compute the jacobian is described in section \ref{sec:gradient} and \ref{sec:lr}
		\ENDFOR
		
		\STATE // Update hidden variables and fit GBDTs from scratch
		\FOR{$i=L$ {\bfseries to} $1$}
		\STATE $\mathbf{h}_i \leftarrow \mathbf{h}_i - \alpha \mathbf{g}_i$
		\STATE $\mathcal{T}_{i-1}, \hat{\mathbf{h}} \leftarrow \{ \}, 0$ // Initialize to an empty list
		\FOR{$j=1$ {\bfseries to} $N_i$}
		\STATE $\mathbf{r} \leftarrow \hat{\mathbf{h}} - \mathbf{h}_i$ // Gradient of MSE
		\STATE $\mathcal{M} \leftarrow RegressionTree(\mathbf{h}_{i-1}, -\mathbf{r})$ // Fit regression tree with variables of previous layer and negative gradient
		\STATE $\hat{\mathbf{h}} \leftarrow \hat{\mathbf{h}} + \beta \mathcal{M}(\mathbf{h}_{i-1})$
		\STATE Append $\mathcal{M}$ to $\mathcal{T}_{i-1}$
		\ENDFOR
		\ENDFOR
		
		\UNTIL{convergence}
		\STATE {\bfseries Return:} $\{\mathcal{T}_0, \mathcal{T}_1,\dots \mathcal{T}_{L-1}\}$
	\end{algorithmic}
\end{algorithm*}

\subsection{Gradient of GBDT Layer}
\label{sec:gradient}
We compute the jacobian of two adjacent GBDT layers to propagate the gradient as follows. We first formalize a single regression tree. Here, it is not our interest how the tree structure is determined. We suppose the tree structure is fixed. Denote $\mathcal{M}: \mathbb{R}^n \rightarrow \mathbb{R}$ as the function of a tree as follows:
\begin{equation}
\mathcal{M}(\mathbf{x}) = \sum_{i \in leaf} \mathcal{I}(\mathbf{x} \in i) f_{\mathbf{w}_i}(\mathbf{x})
\end{equation}
where $leaf$ is a set of leaves, $\mathcal{I}$ is the indicator function and $f_{\mathbf{w}_i}$ is the function of leaf $i$. That is, we first decide which leaf an input belongs to, then we compute the prediction of that leaf. The gradient of $\mathcal{M}$ w.r.t its input is as follows:
\begin{equation}
\D{\mathcal{M}(\mathbf{x})}{\mathbf{x}} = \sum_{i \in leaf} \mathcal{I}(\mathbf{x} \in i) \D{f_{\mathbf{w}_i}(\mathbf{x})}{\mathbf{x}} + \D{\mathcal{I}(\mathbf{x} \in i)}{\mathbf{x}} f_{\mathbf{w}_i}(\mathbf{x})
\end{equation}

The indicator function should be fixed if we suppose the tree structure is fixed. This means $\D{\mathcal{I}(\mathbf{x} \in i)}{\mathbf{x}} = 0, \forall i \in leaf$. That is, an individual sample has no effect on the tree structure. This is not true but it is a good approximation as the training samples become massive. Then, the gradient is simplified as:
\begin{equation}
\D{\mathcal{M}(\mathbf{x})}{\mathbf{x}} = \sum_{i \in leaf} \mathcal{I}(\mathbf{x} \in i) \D{f_{\mathbf{w}_i}(\mathbf{x})}{\mathbf{x}}
\end{equation}

Note that if $f$ is a constant function, the gradient degrades to 0. In this case, we are not able to learn multi-layered GBDT via BP. This is why we replace the constant value with linear regression. We have derived the gradient of an individual regression tree. Since the prediction of GBDT is the summation of the predictions of all regression trees, the gradient of GBDT is easily obtained. Denote $\mathcal{T}: \mathbb{R}^n \rightarrow \mathbb{R}$ as the function of GBDT. Then, we get:
\begin{equation}
\D{\mathcal{T}}{\mathbf{x}} = \sum_{i} \D{\mathcal{M}_i}{\mathbf{x}}
\end{equation}

Suppose given $\mathbf{h}_{i-1}$, $j$-th element of $\mathbf{h}_i$ is predicted by $j$-th GBDT, i.e. $\mathcal{T}^j$. Then, the jacobian is as follows:
\begin{equation}
\D{\mathbf{h}_i}{\mathbf{h}_{i-1}} = \left( \D{\mathcal{T}^0}{\mathbf{h}_{i-1}}, \D{\mathcal{T}^1}{\mathbf{h}_{i-1}}, \dots \D{\mathcal{T}^{d_i}}{\mathbf{h}_{i-1}} \right)^T
\end{equation}
Once we get the jacobian, we can propagate the gradient of multi-layered GBDT using \eqref{eq_bp_h}.

\subsection{Linear Regression at Leaves}
\label{sec:lr}

\begin{figure}
	\centering
	\includegraphics[width=0.45\textwidth]{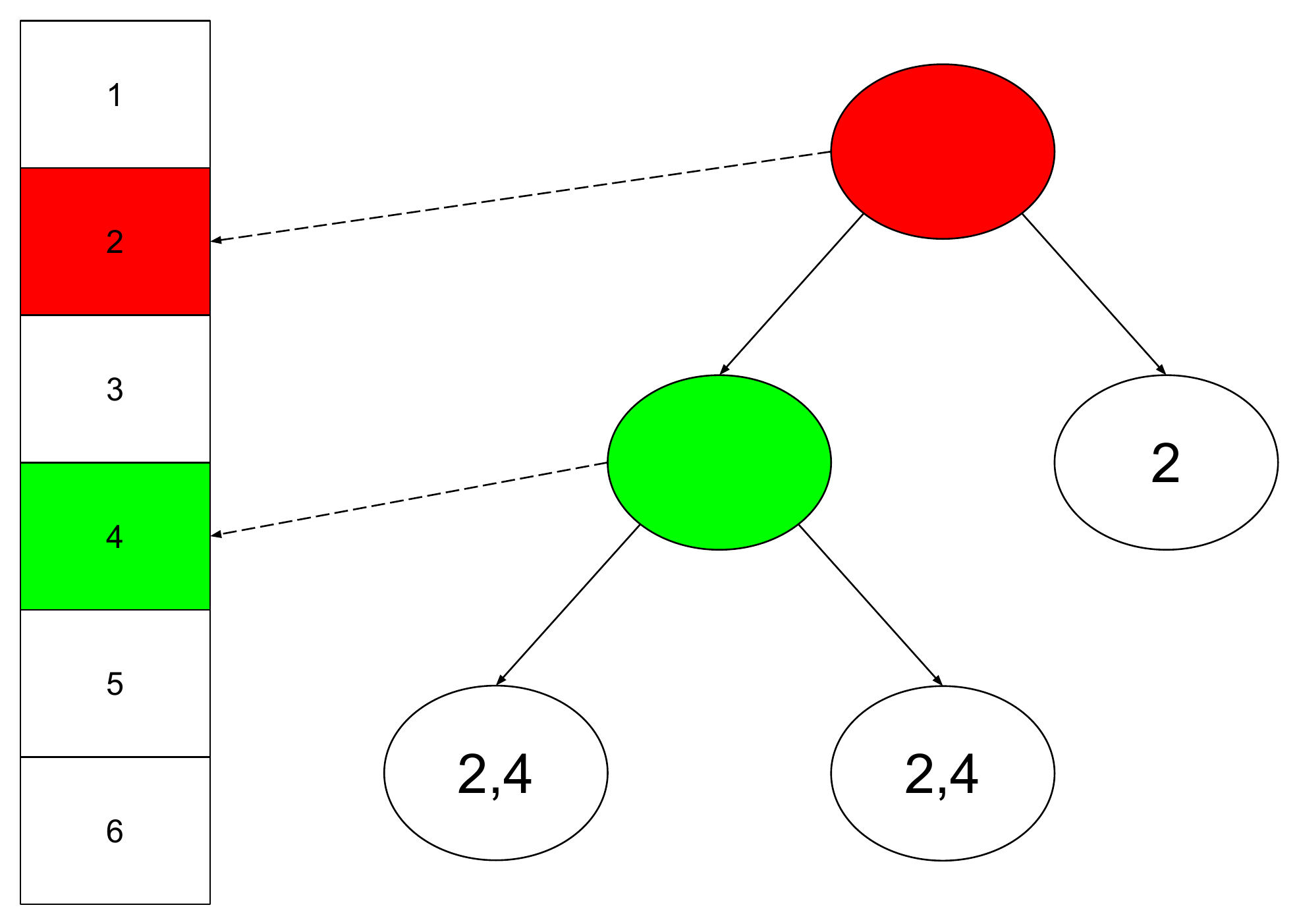}
	\caption{Example of incremental feature selection. Each circle is a leaf, while numbers in a circle are the selected features for that leaf.}
	\label{fig:select}
\end{figure}

The jacobian depends on leaf functions, and we describe how to learn leaf functions. As mentioned earlier, the parameters of each leaf are learned by linear regression. In \cite{de2017linxgboost}, all dimensions of inputs are used for linear regression. However, the drawbacks of such a strategy are obvious, as addressed in \cite{shi2018gradient}. First, the computational cost is relatively high. Second, given samples in a leaf, not all dimensions are useful for predicting their targets, thus such a strategy has a high risk of over-fitting. Feature selection is required. In this work, we follow the incremental feature selection proposed in \cite{shi2018gradient}. Given a leaf, we trace back its decision path. For each node in the decision path, there is a dimension used to split. We collect all splitting dimensions that appeared in the decision path. Then, we only use those dimensions to learn the linear regression model. As shown in Fig.~\ref{fig:select}, the number of selected dimensions is no more than the depth of trees. Those dimensions are relevant to the targets. Another useful trick for linear regression is extending the feature dimensions based on some transforms, which increases the capacity of linear regression.

Denote $\mathbf{x}\in \mathbb{R}^d$ as a sample with $d$ dimensions. Here, we suppose these $d$ dimensions are already captured by incremental feature selection. Denote $\mathbf{X} \in \mathbb{R}^{n\times d}$ as a collection of $n$ samples belonging to the same leaf. Denote $\mathbf{y}\in \mathbb{R}^n$ as their current targets (negative residuals). Denote $\phi$ as the feature extending function. Then, the objective of linear regression is as follows:
\begin{equation}
\label{eq:lr}
\Arrowvert \phi(\mathbf{X})\mathbf{w} - \mathbf{y} \Arrowvert_2^2 + \frac{\lambda}{2}\mathbf{w}^T\mathbf{w}
\end{equation}
where $\lambda$ is the strength of $L_2$ regularization. The solution of~\eqref{eq:lr} is as follows:
\begin{equation}
\mathbf{w}^* = [\phi(\mathbf{X})^T\phi(\mathbf{X}) + \lambda\mathbf{I}]^{-1}\phi(\mathbf{X})^T\mathbf{y}
\end{equation}
Then, the predictions of this leaf is as follows:
\begin{equation}
f_{\mathbf{w}^*}(\mathbf{x}) = \phi(\mathbf{x})^T\mathbf{w}^*
\end{equation}
The gradient of the leaf function is as follows:
\begin{equation}
\D{f_{\mathbf{w}^*}(\mathbf{x})}{\mathbf{x}} = \left( \D{\phi(\mathbf{x})}{\mathbf{x}} \right)^T\mathbf{w}^*
\end{equation}
It is easy to choose a differentiable $\phi$. From Sections~\ref{sec:gradient} and~\ref{sec:lr}, we get the jacobian of GBDT layers and thus back propagation is operational.

\subsection{Implementation Details}
For each regression tree, we use $DecisionTreeRegressor$ in scikit-learn package \cite{pedregosa2011scikit} which splits a node based on ``friedman mse''. Although recent second order based implementations show better performance, there is no significant difference in our case because MSE is used as the loss function to fit hidden variables. Although \cite{de2017linxgboost} developed a novel splitting objective for linear regression at leaves, we still use ``friedman mse'' for a fair comparison and easy implementation. The splitting objective is not the goal of this work. For feature extending function, we set $\phi$ as follows:
\begin{equation}
\label{eq:trans}
\phi(\mathbf{x}) = [1, \mathbf{x}, \mathbf{x}^2]
\end{equation}
Other choices for feature extending can be possible.

Momentum is commonly used when training neural networks with stochastic gradient descent to speed up the learning process. We introduce this technique into GBDT-BP as follows. Denote $\mathbf{m}_i$ as the momentum of $\mathbf{h}_i$ as follows:
\begin{equation}
\mathbf{m}_i \leftarrow \mu\mathbf{m}_i + (1-\mu)\D{\mathcal{L}}{\mathbf{h}_i}
\end{equation}
When we update $\mathbf{h}_i$, we replace its gradient with $\mathbf{m}_i$ in~\eqref{eq:update}. We set $\mu=0.5$ and $\alpha=0.5$ in all experiments. Hidden variable $\mathbf{h}$ is initialized as random Gaussian noise.

\begin{figure*}[t]
	\centering
	\subfigure[Circle]{
		\label{fig:circle_o}
		\includegraphics[width=0.23\textwidth]{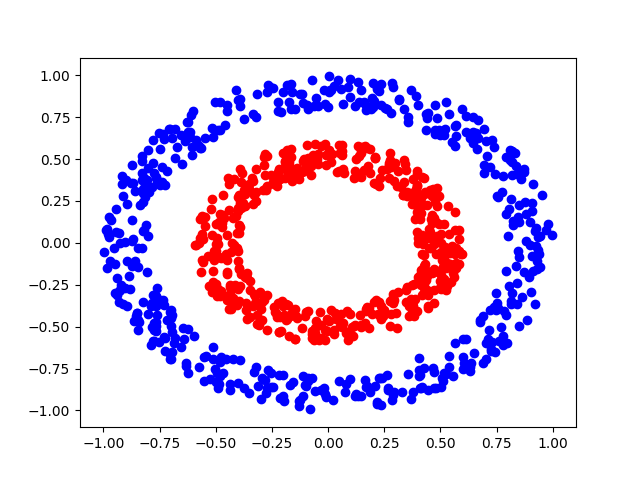}}
	\subfigure[Epoch $2$]{
		\includegraphics[width=0.23\textwidth]{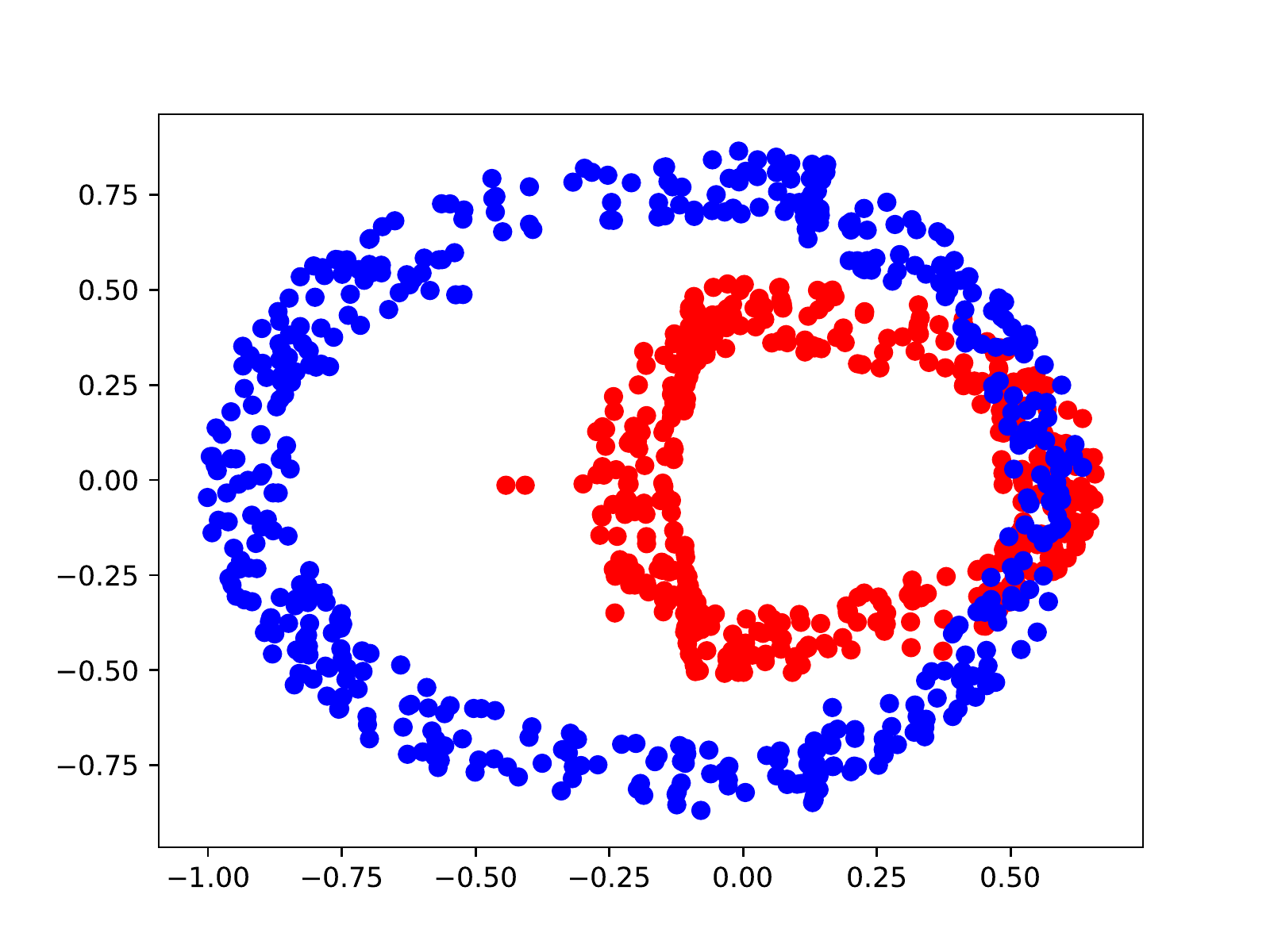}}
	\subfigure[Epoch $4$]{
		\includegraphics[width=0.23\textwidth]{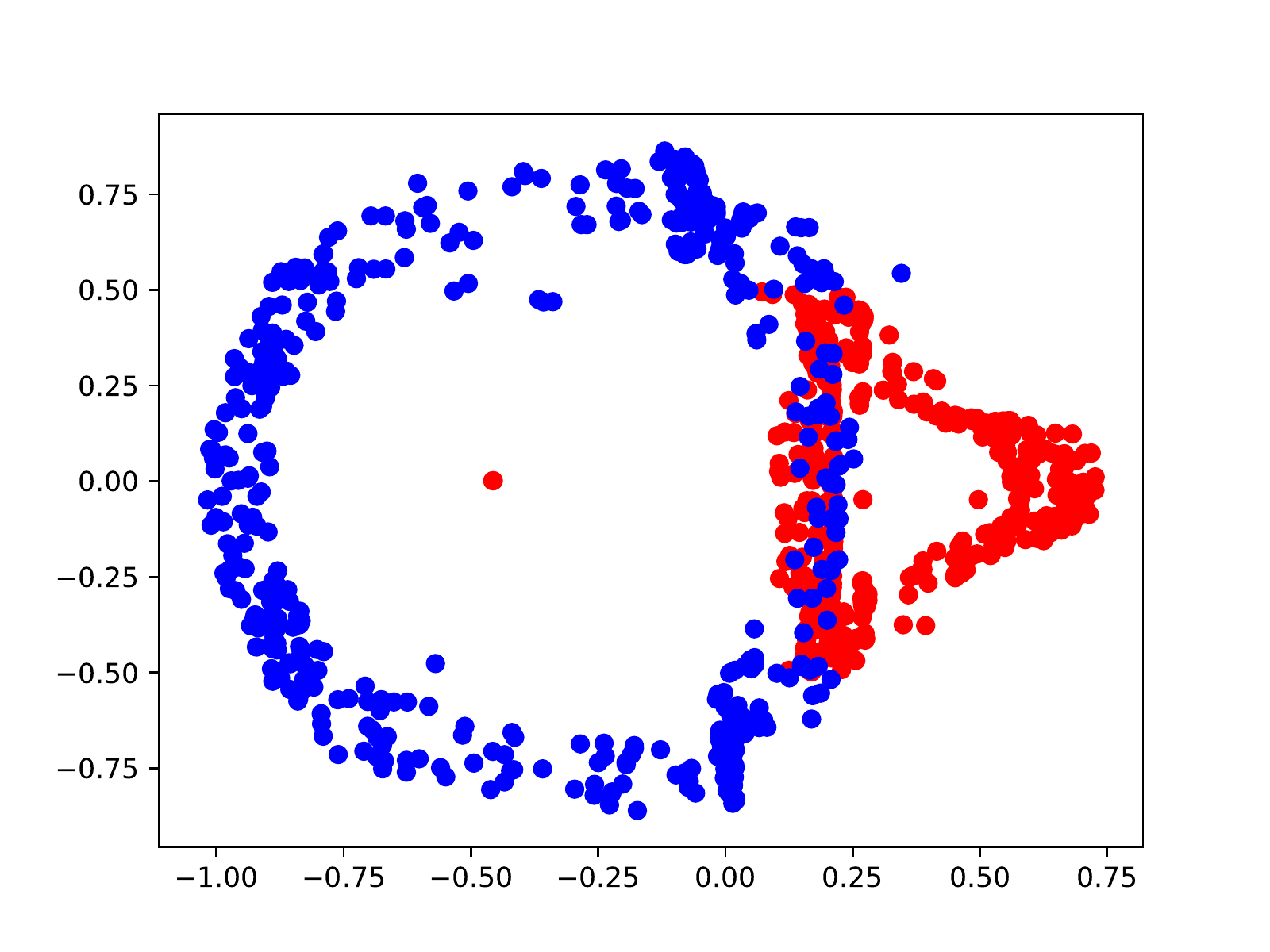}}
	\subfigure[Epoch $6$]{
		\includegraphics[width=0.23\textwidth]{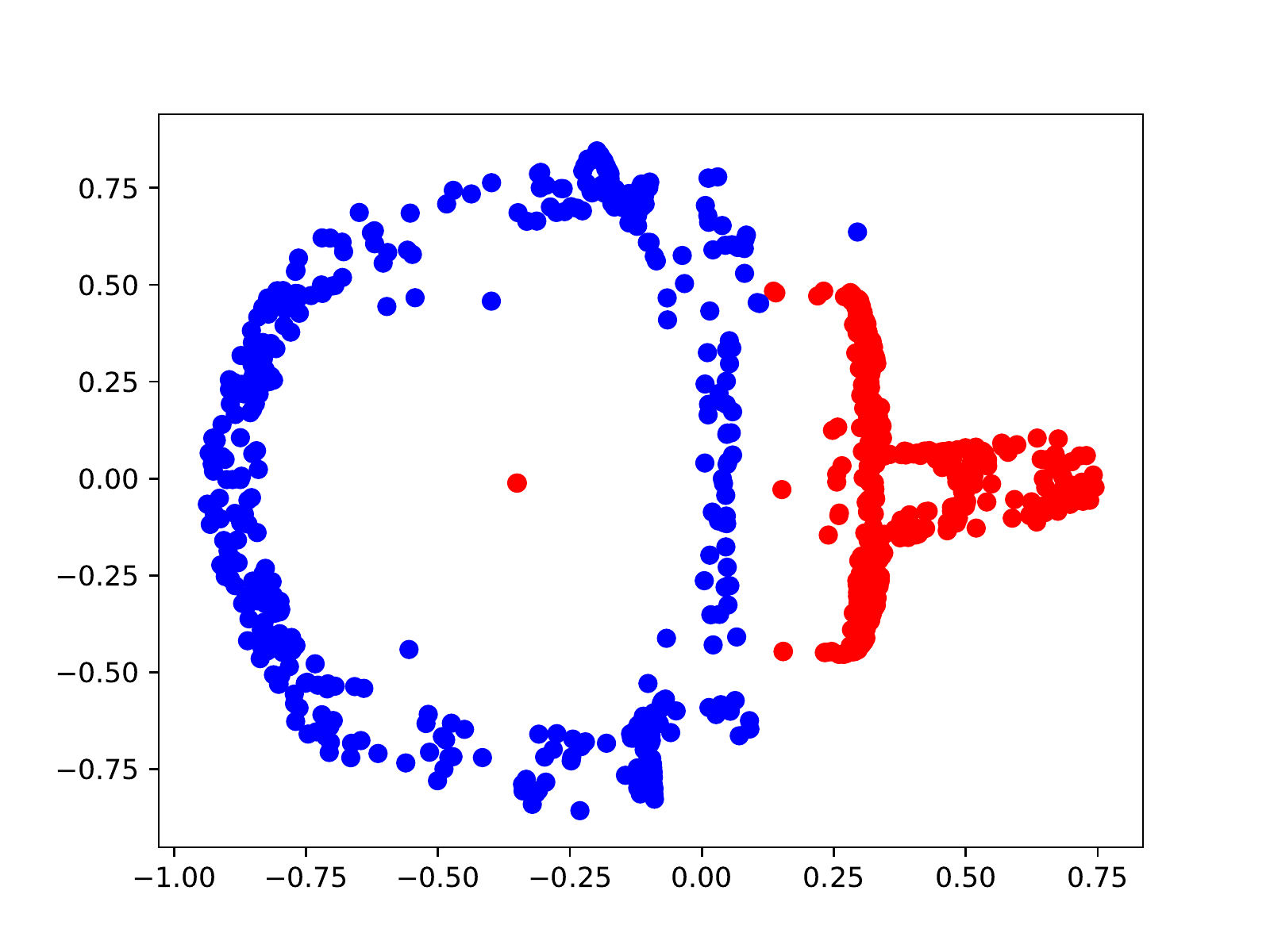}}
	\subfigure[Epoch $11$]{
		\includegraphics[width=0.23\textwidth]{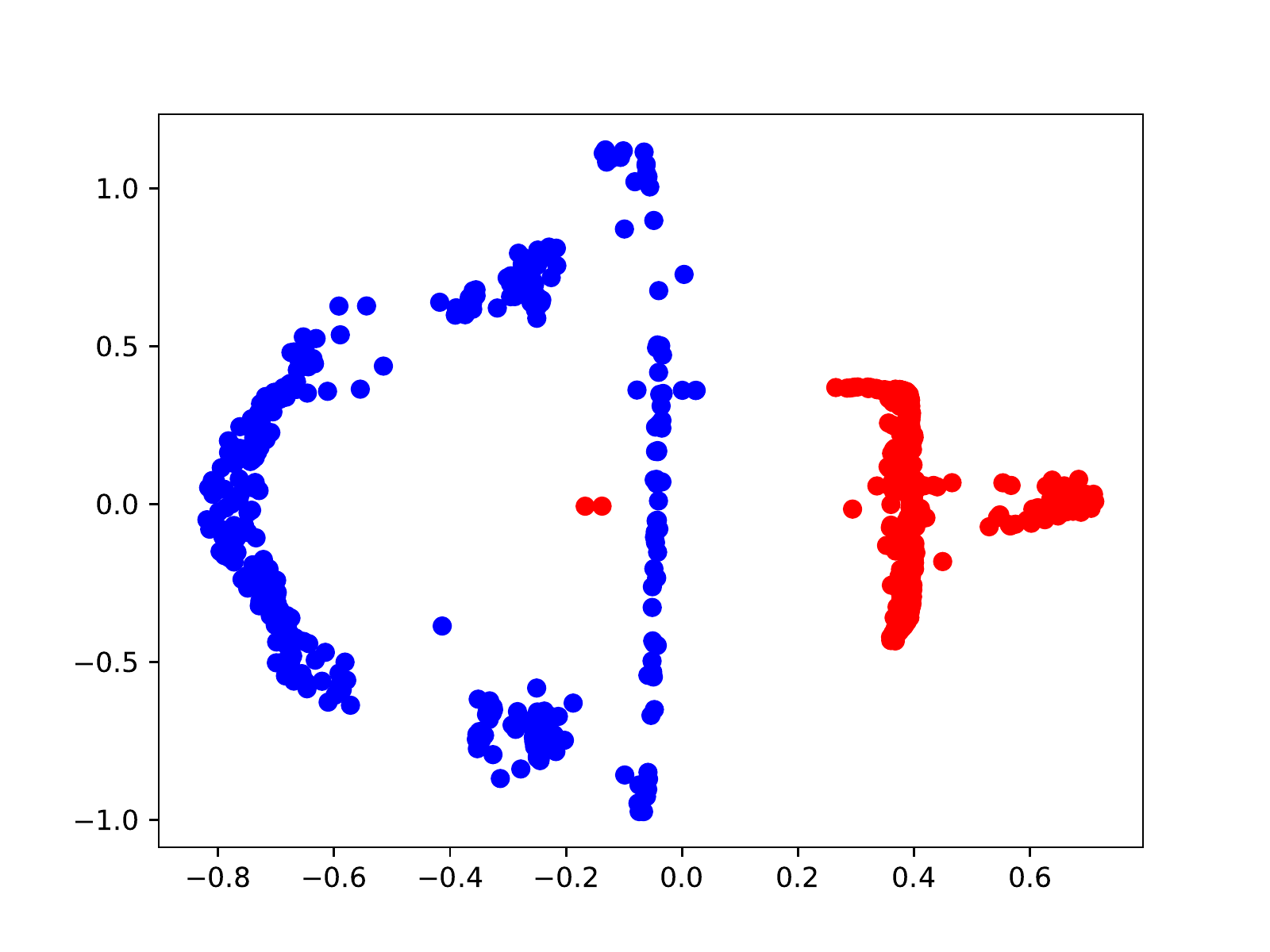}}
	\subfigure[Epoch $13$]{
		\includegraphics[width=0.23\textwidth]{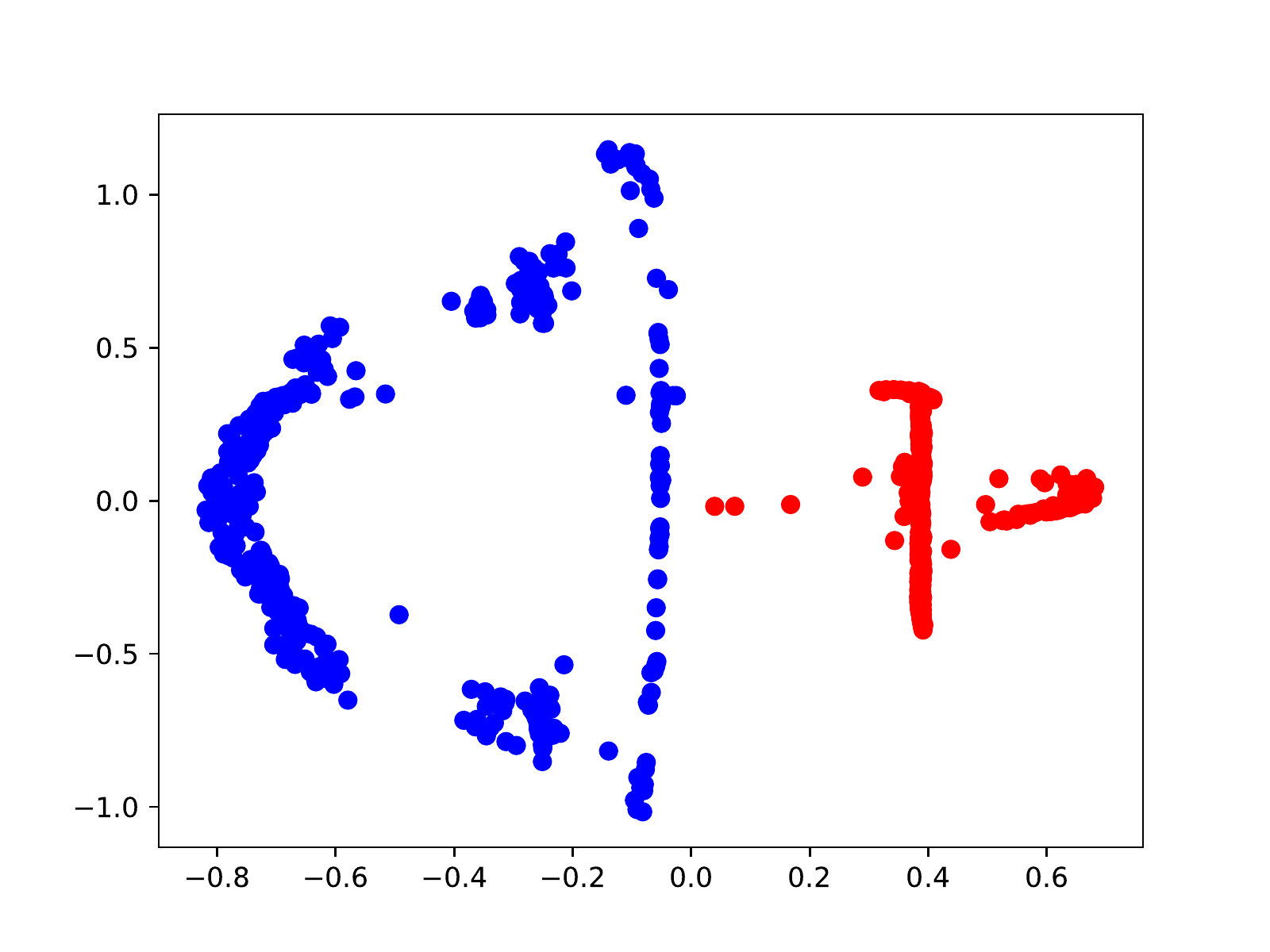}}
	\subfigure[Epoch $17$]{
		\includegraphics[width=0.23\textwidth]{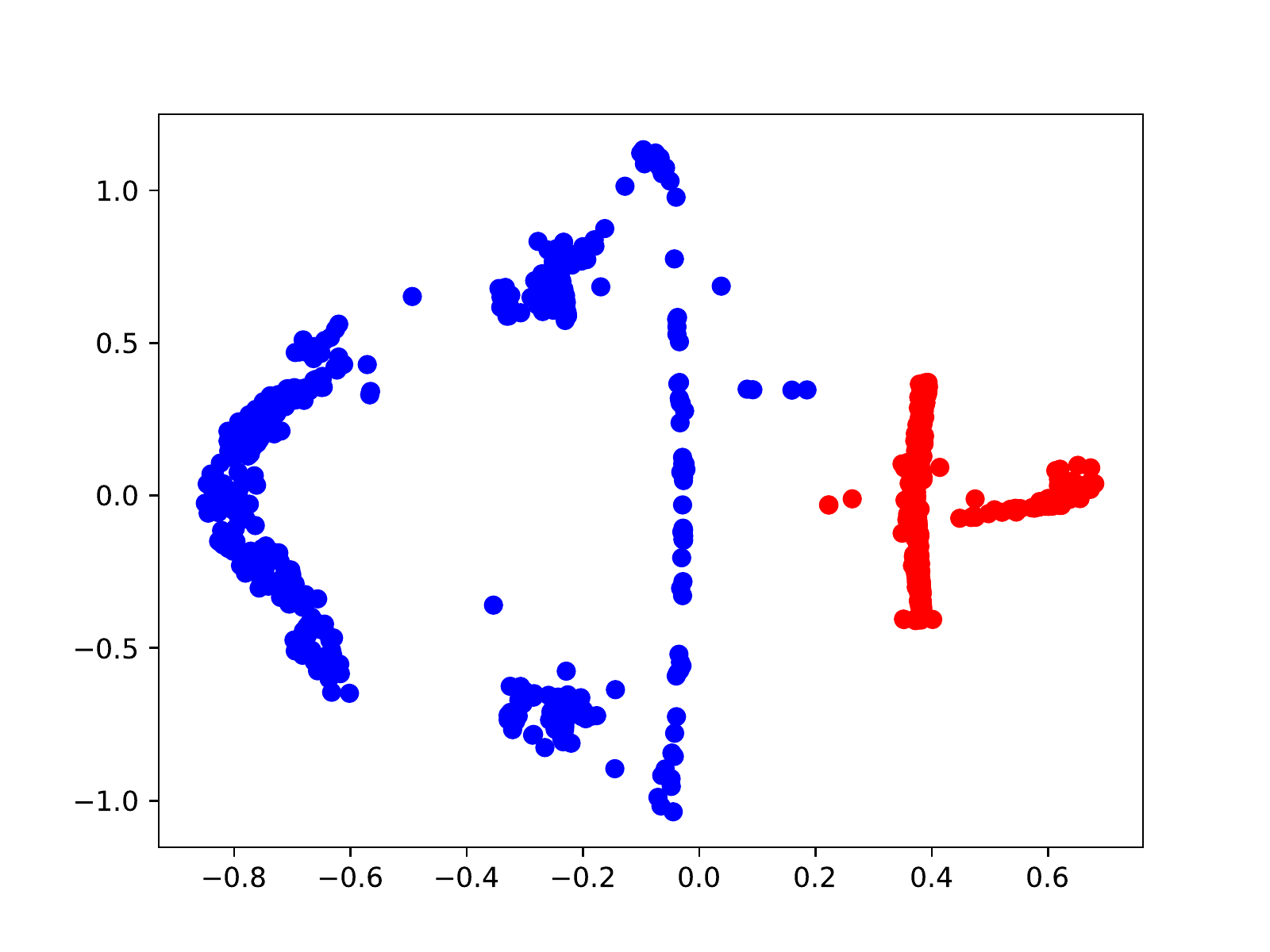}}
	\subfigure[Epoch $23$]{
		\includegraphics[width=0.23\textwidth]{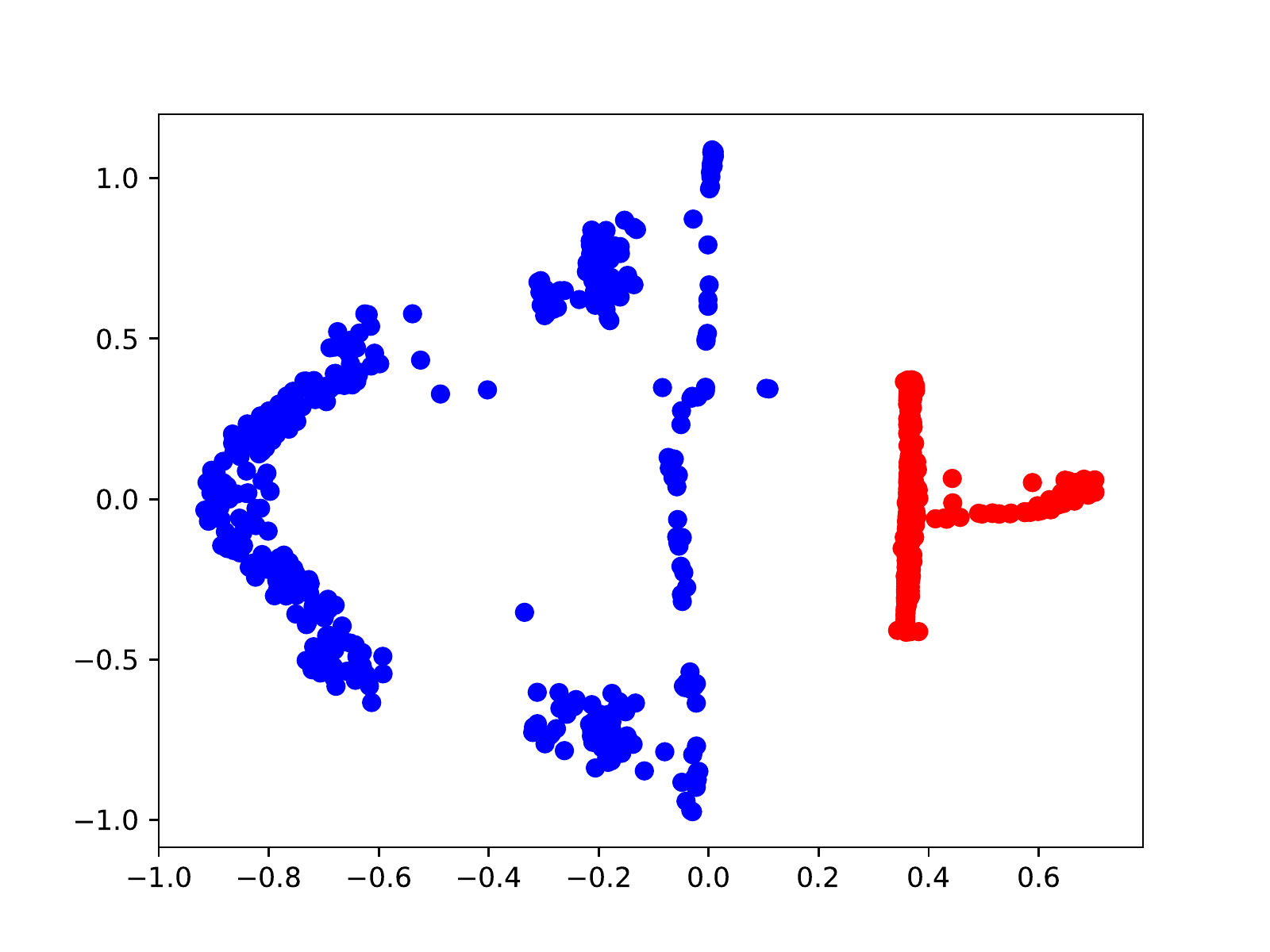}}
	\caption{Progress of hidden representations on $Circle$. We train a 2-2-2 multi-layered GBDT for binary classification to learn meaningful hidden representations.}
	\label{fig:circle}
\end{figure*}

\begin{figure*}[ht]
	\centering
	\subfigure[Curve]{
		\label{fig:curve_o}
		\includegraphics[width=0.24\textwidth]{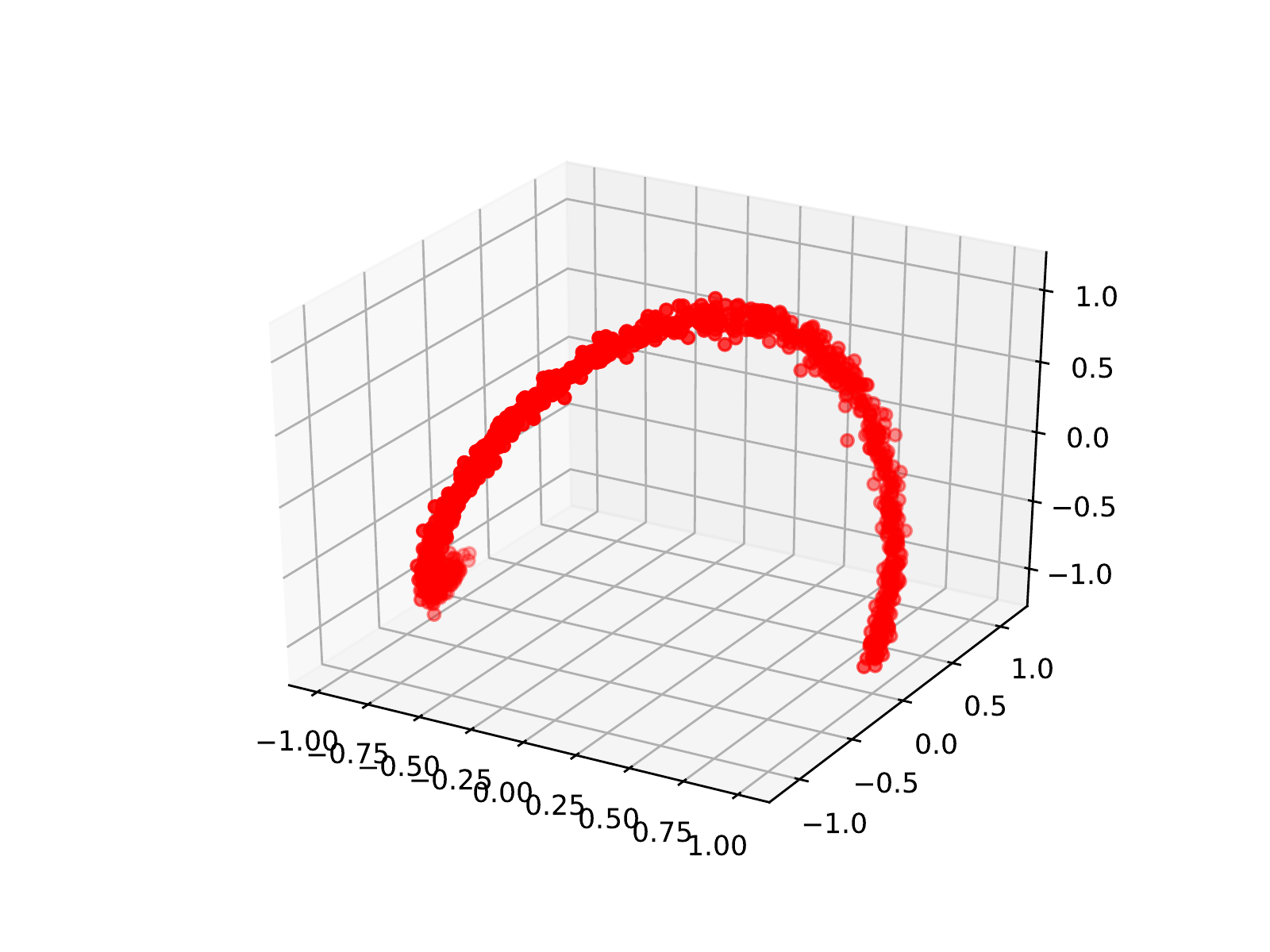}}
	\subfigure[Epoch $2$]{
		\includegraphics[width=0.24\textwidth]{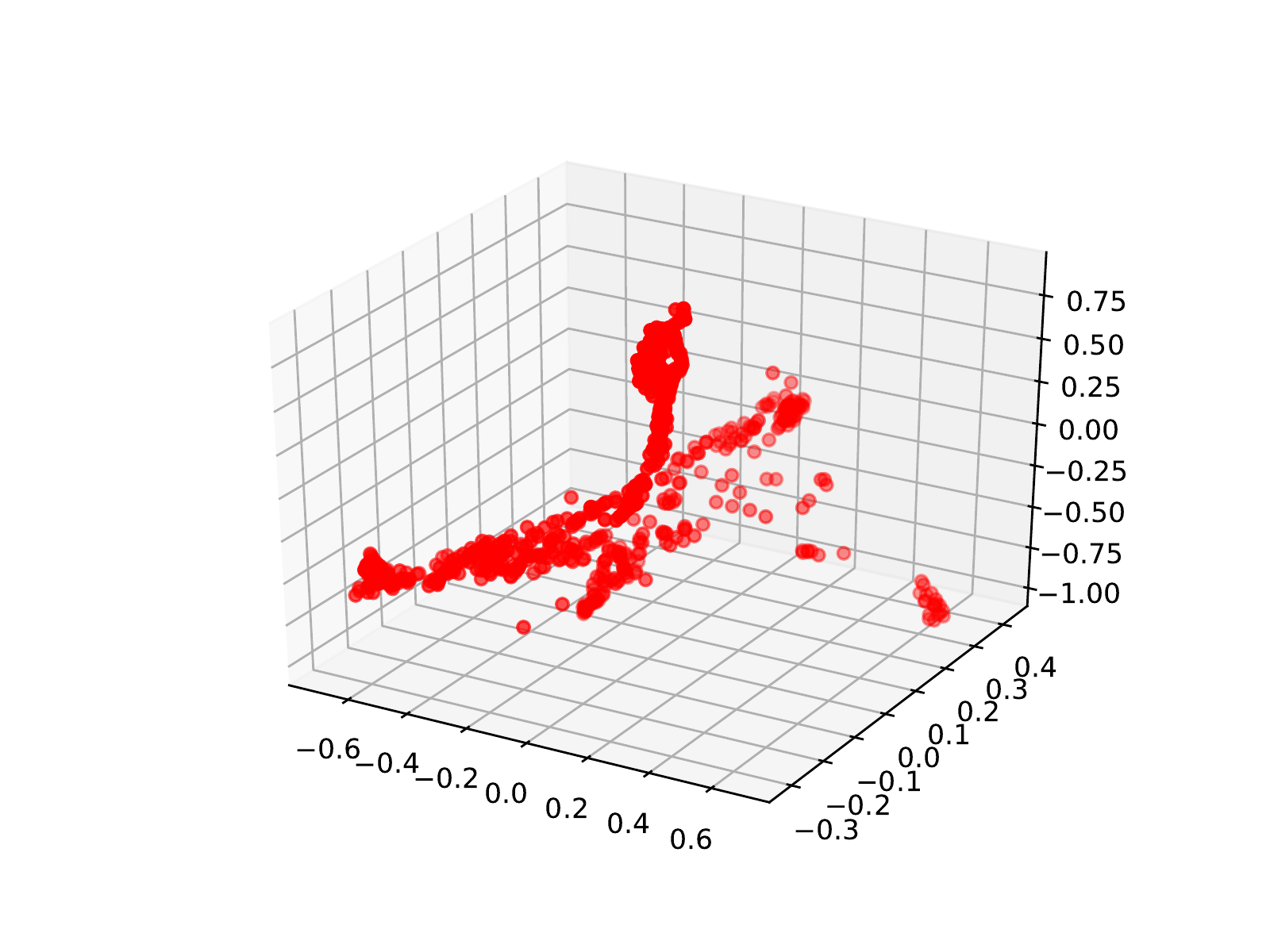}}
	\subfigure[Epoch $4$]{
		\includegraphics[width=0.24\textwidth]{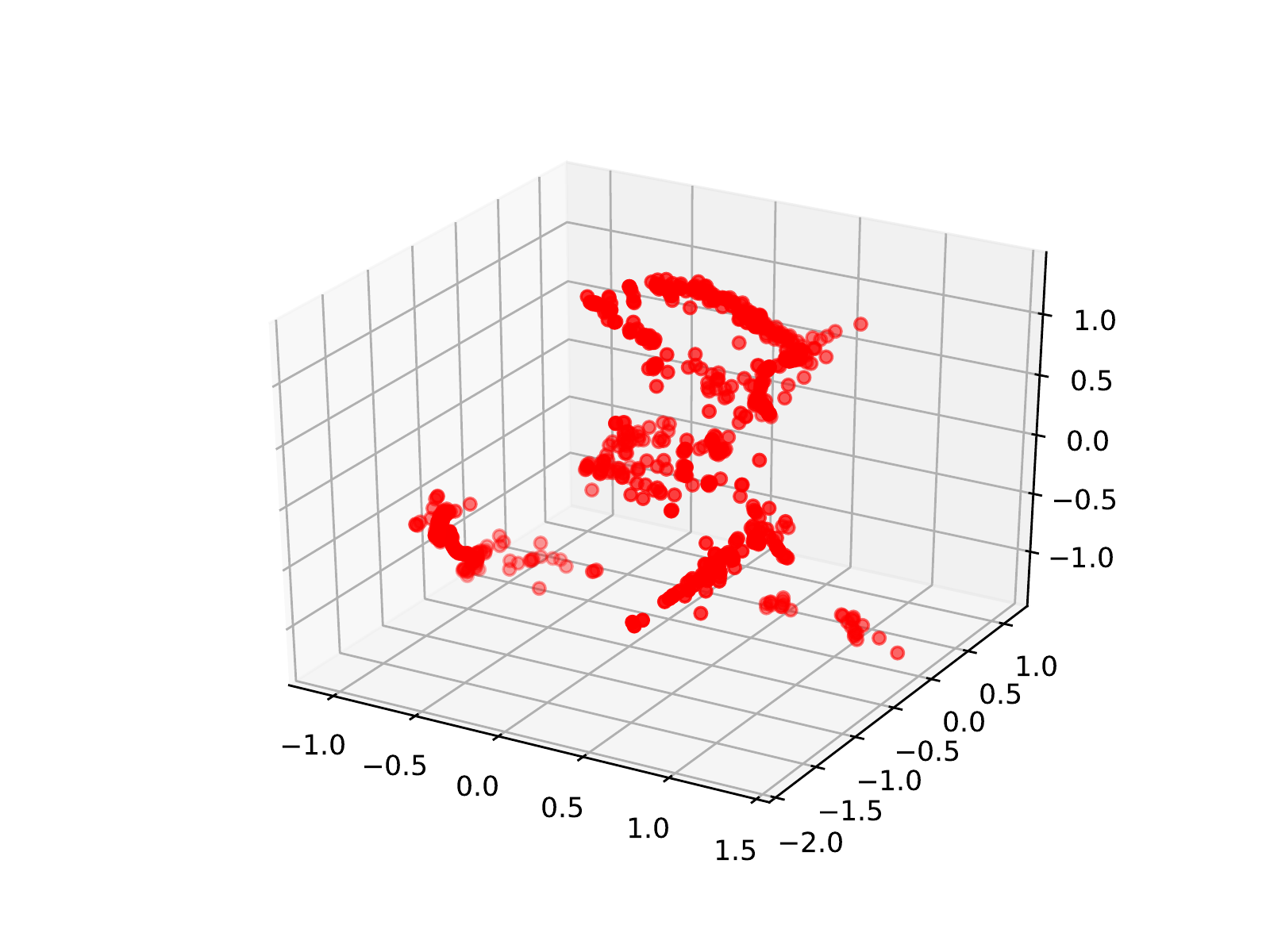}}
	\subfigure[Epoch $6$]{
		\includegraphics[width=0.24\textwidth]{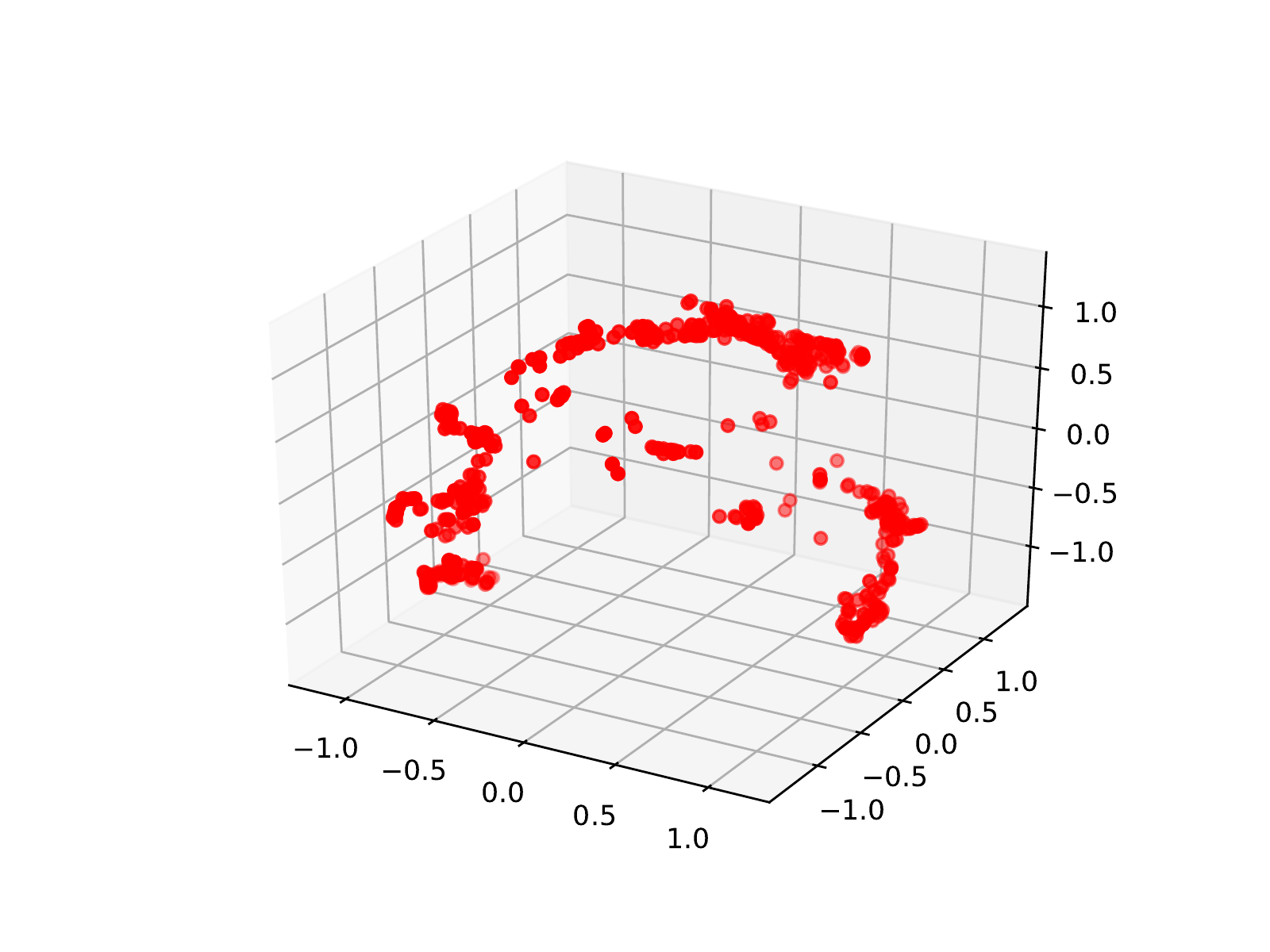}}
	\subfigure[Epoch $10$]{
		\includegraphics[width=0.24\textwidth]{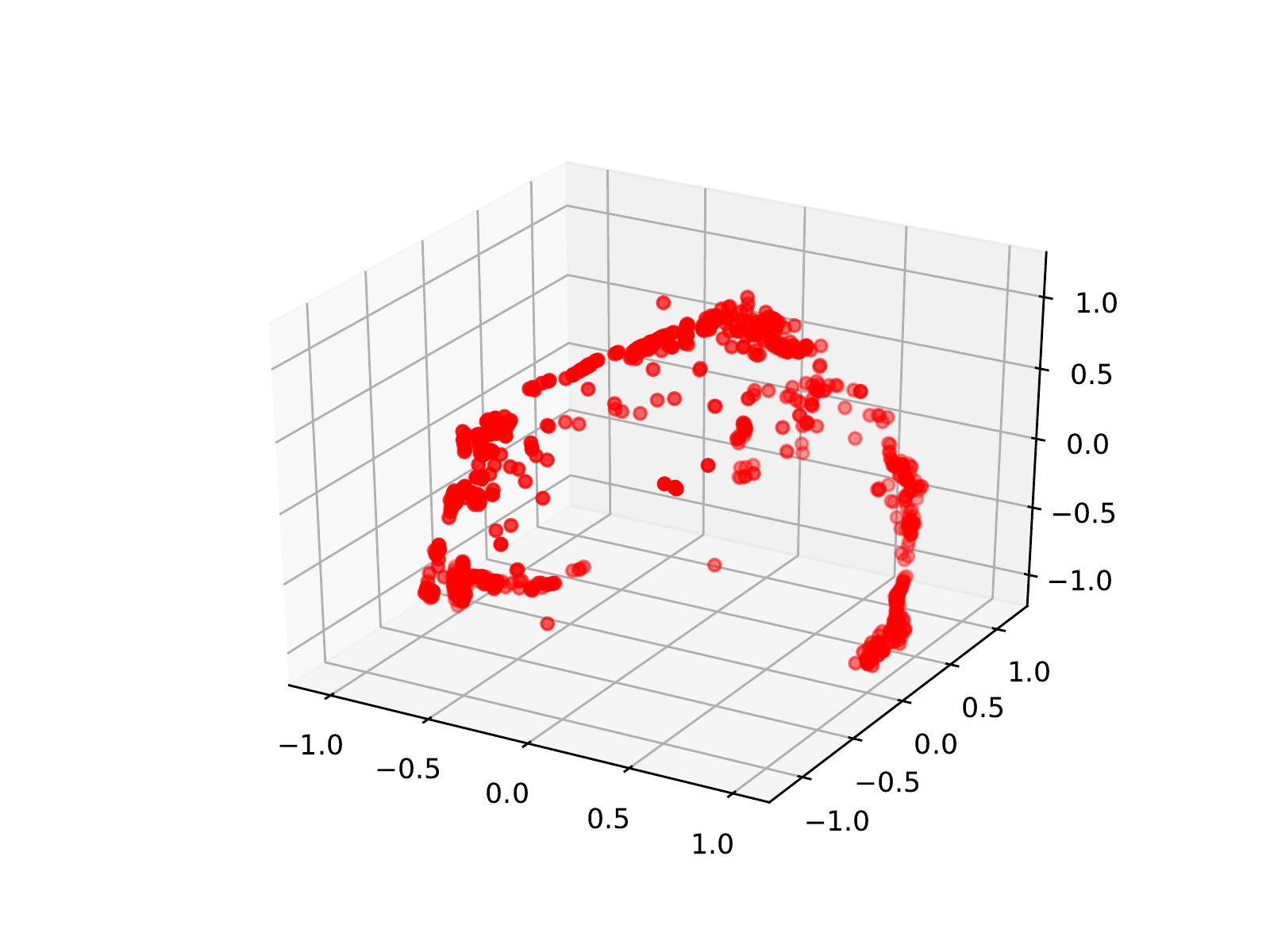}}
	\subfigure[Epoch $15$]{
		\includegraphics[width=0.24\textwidth]{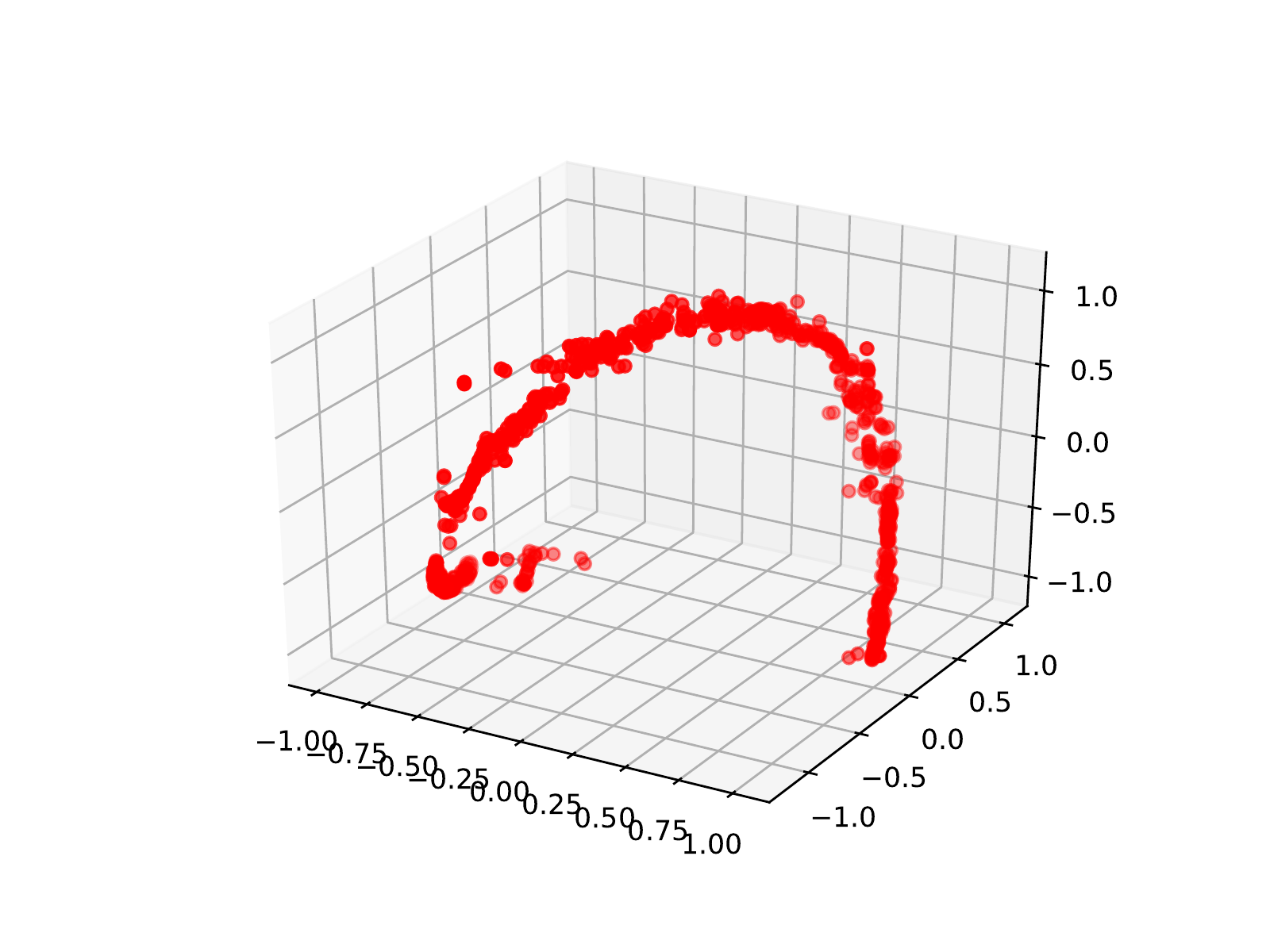}}
	\subfigure[Epoch $25$]{
		\includegraphics[width=0.24\textwidth]{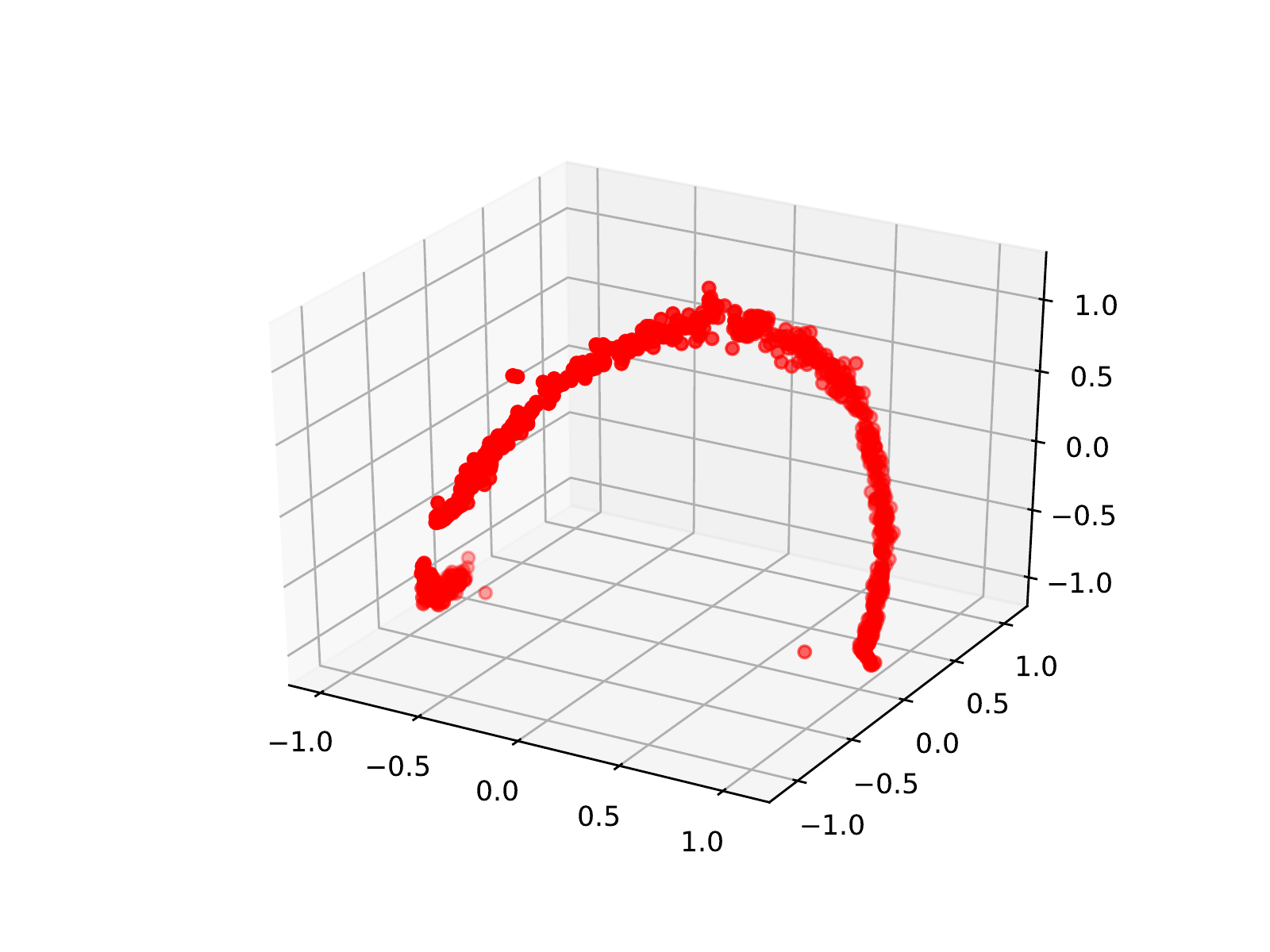}}
	\subfigure[Epoch $47$]{
		\includegraphics[width=0.24\textwidth]{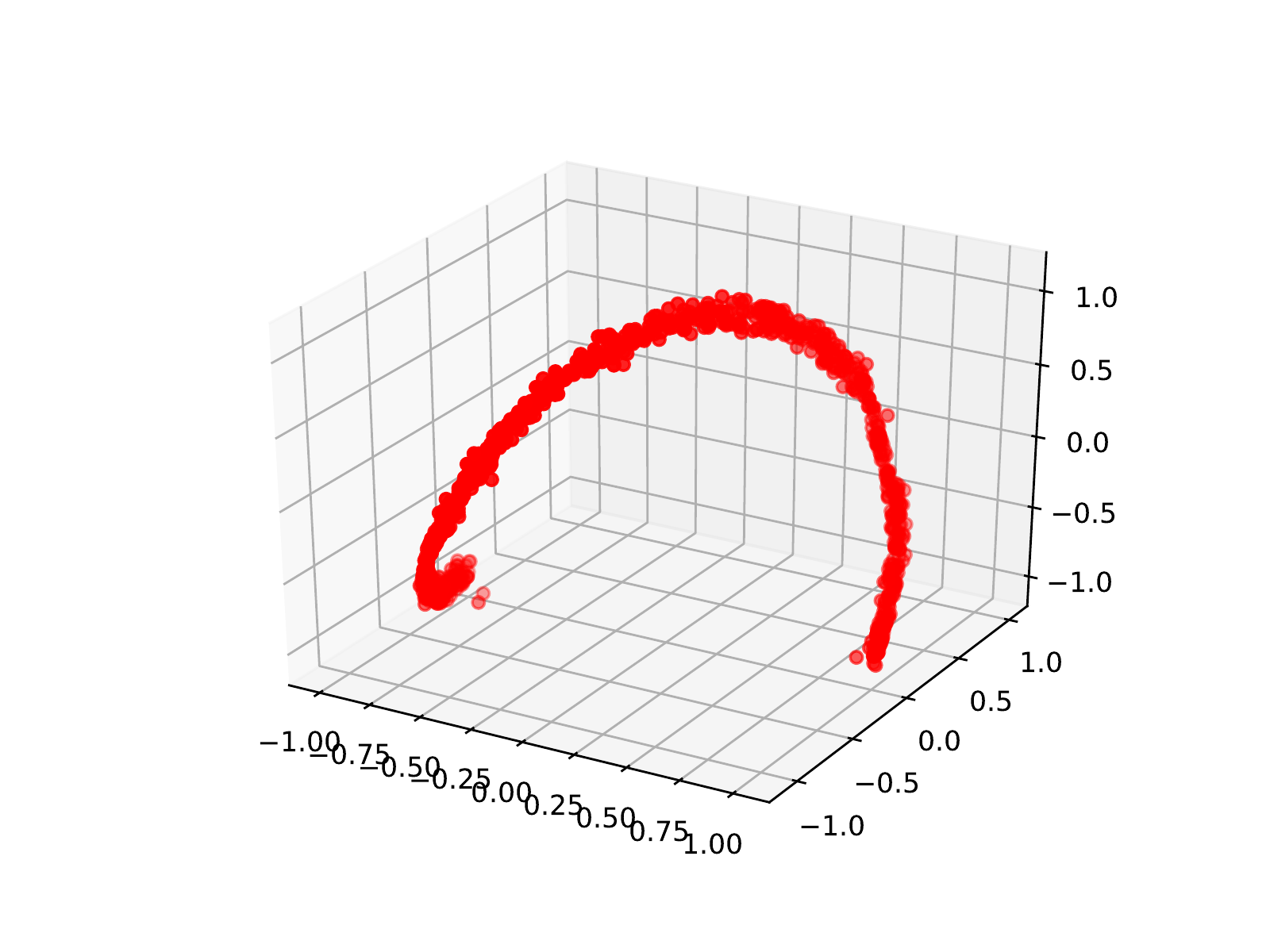}}
	\caption{Progress of final reconstructions on $Curve$. We train a 3-2-3 multi-layered GBDT as an auto-encoder to learn the 2D manifold of 3D data.}
	\label{fig:curve}
\end{figure*}

\section{Experiments}

\begin{table*}[t]
	\caption{RMSE comparison for fitting a random neural network function.}
	\label{tab:rand}
	\begin{center}
		\begin{tabular}{lcccr}
			\toprule
			Method & RMSE & Constant Leaf & Multi Layer \\
			\midrule
			GBDT    & 0.01867 $\pm$ 0.00023 & $\surd$& $\times$ \\
			XGBoost & 0.01886 $\pm$ 0.00018 & $\surd$& $\times$\\
			GBDT-PL    & 0.01861 $\pm$ 0.00022& $\times$& $\times$ \\
			\hline
			GBDT-BP    & \textbf{0.01775 $\pm$ 0.00026}& $\times$& $\surd$ \\
			\bottomrule
		\end{tabular}
	\end{center}
\end{table*}

\begin{table*}[t]
	\caption{Classification accuracy on \textit{Income Prediction} and \textit{Protein Localization} datasets. NN is fully connected neural networks, while $^*$ indicates the result is reported from literature \cite{feng2018multi}.  }
	\label{tab:tab}
	\begin{center}
		\begin{tabular}{lllll}
			\toprule
			Method &  \textit{Income Prediction} & \textit{Protein Localization} & \textit{Drug Consumption} & \textit{MNIST-3000}\\
			\midrule
			XGBoost     & 0.8721 & 0.5951 $\pm$ 0.0341& 0.7040 $\pm$ 0.0251 & 0.9242\\
			GBDT-MO   & 0.8730 & 0.6079 $\pm$ 0.0303& 0.7047 $\pm$ 0.0237 & 0.9263\\
			GBDT-PL     & 0.8726 & 0.6037 $\pm$ 0.0308 & 0.6962 $\pm$ 0.0265 & 0.9254\\
			NN              & 0.8543$^*$ & 0.5907 $\pm$ 0.0268$^*$ & 0.6954 $\pm$ 0.0291 & 0.9305\\
			mGBDT        & \textbf{0.8742}$^*$ & 0.6160 $\pm$ 0.0323$^*$ & 0.6990 $\pm$ 0.0321 & 0.9284\\
			\hline
			GBDT-BP    & 0.8741& \textbf{0.6186 $\pm$ 0.0296} & \textbf{0.7063 $\pm$ 0.0306} & \textbf{0.9317}\\
			\bottomrule
		\end{tabular}
	\end{center}
\end{table*}

\begin{table*}[thb]
	\caption{Classification accuracy on \textit{Protein Localization} dataset with varying number of layers. The dimension of all hidden layers is 16. NN is fully connected neural networks, while $^*$ indicates that the result is reported from literature \cite{feng2018multi}.}
	\label{tab:layer}
	\begin{center}
		\begin{tabular}{lcccr}
			\toprule
			Hidden Layer &  NN$^*$ & mGBDT$^*$ & GBDT-BP \\
			\midrule
			1   & 0.5803 $\pm$ 0.0316 & 0.6160 $\pm$ 0.0323 & 0.6186 $\pm$ 0.0296\\
			2     & 0.5907 $\pm$ 0.0268 & 0.5948 $\pm$ 0.0268 & 0.5938 $\pm$ 0.0382\\
			3     & 0.5901 $\pm$ 0.0270 & 0.5897 $\pm$ 0.0312 & 0.5882 $\pm$ 0.0461\\
			4     & 0.5768 $\pm$ 0.0286 & 0.5782 $\pm$ 0.0229 & 0.5751 $\pm$ 0.0291\\
			\bottomrule
		\end{tabular}
	\end{center}
\end{table*}

We first evaluate GBDT-BP on two synthetic datasets: \emph{Circle} and \emph{Curve}. We train GBDT-BP on them in supervised setting and unsupervised setting, respectively. Then, we compare the function approximation capability of GBDT-BP by fitting a random neural network.
\hl{Finally, we evaluate GBDT-BP on three tabular datasets and \emph{MNIST}\footnote{http://yann.lecun.com/exdb/mnist/}. All tabular datasets are from UCI \cite{Dua:2017}: \emph{Income Prediction}\footnote{http://archive.ics.uci.edu/ml/datasets/Adult}, \emph{Protein Localization}\footnote{http://archive.ics.uci.edu/ml/datasets/Yeast}, and \emph{Drug Consumption}\footnote{https://archive.ics.uci.edu/ml/datasets/ \\ Drug+consumption+\%28quantified\%29}. \emph{Income Prediction} and \emph{Protein Localization} are also used in \cite{feng2018multi}, and we choose them for a fair comparison. 
}
We obtain the best performance by adjusting the depth of trees and the number of boosters as \cite{feng2018multi}. GBDT-BP is convergent within 60 epochs for all datasets. 

We denote $\mathcal{U}$ is a uniform distribution, while $\mathcal{N}$ is a normal distribution.

\subsection{Synthetic Data}

\emph{Circle}: This dataset is used for binary classification as shown in Fig.~\ref{fig:circle_o}. The angle of each point is sampled from $\mathcal{U}(0, 2\pi)$. The radius of blue points is sampled from $\mathcal{U}(0.8, 1.0)$, while the radius of red points is sampled from $\mathcal{U}(0.4, 0.6)$. In this way, we generate $10000$ points for training. We build a single hidden layer GBDT-BP model with structure $2\rightarrow2\rightarrow1$. The maximum tree depth is set to 6 and the number of boosters per GBDT is set to 8. We show how hidden representations progress in Fig.~\ref{fig:circle}. It is obvious that the learned hidden representations make the classification easier. At the end of the progress, we can completely separate them by a single node of a decision tree.

\emph{Curve}: This dataset is used for self-reconstruction as shown in Fig.~\ref{fig:curve_o}. Each 3D point is sampled by:
\begin{equation}
[t, \mathcal{N}(0, 0.05) + \sin t, \mathcal{N}(0, 0.05) + \cos t]
\end{equation}
where $t$ is $\mathcal{U}(-1, 1)$. Those points belong to a 2D manifold.

We generate $10000$ points for training. We train a GBDT-BP auto-encoder with structure $3\rightarrow2\rightarrow3$. The maximum tree depth is set to 5 and the number of boosters per GBDT is set to 32. We provide how final reconstructions progress in Fig.~\ref{fig:curve}. As shown in the figure, the 2D manifold is successfully captured. Those results indicate that GBDT-BP is feasible in both supervised setting and unsupervised setting. Moreover, it learns meaningful hidden representations of data.

\emph{Random neural network}: We check the function approximation capability of GBDT-BP by fitting it to a random neural network function. Specifically, the input $\mathbf{x} \in \mathbb{R}^{32}$ is sampled from $\mathcal{U}(0, 1)$. Its corresponding target is computed by the following random initialized neural network:
\begin{equation}
Dense(32, 16) \rightarrow ReLU \rightarrow Dense(16, 1) \nonumber
\end{equation}
We generate 6000 pairs of samples, and split them into training and test sets by 10-fold cross-validation\footnote{We randomly split the samples into 10 folds. At each time, we use one of those folds as test set, while we use the others as training set. We repeat this process 10 times until all folds are used as test set. We report mean and standard deviation of the results.}. We compare GBDT-BP with GBDT, XGBoost and GBDT-PL. Here, GBDT-PL is implemented by us as the building block of GBDT-BP. We measure the performance in terms of root mean square error (RMSE). The GBDT-BP model has a single hidden layer with 32 variables. The numbers of boosters are 40 and 200 for the first and second layer. For single-layered models, we select the best RMSE within 2000 boosters. GBDT-BP is significantly better than others. From Table~\ref{tab:rand}, it can be observed that linear regression at leaves has no remarkable effects on the performance. Thus, it can be concluded that the performance gain is contributed by multi-layered representations.

\subsection{Real Data}
\hl{
	For real data, we compare GBDT-BP with GBDT-PL, XGBoost, fully connected neural networks (NN) trained with back propagation, and mGBDT. Experimental results are presented in Table~\ref{tab:tab}. On every dataset, we keep the experimental settings as same for all methods.
}

\emph{Income Prediction}: This dataset consists of 32561 training samples and 16281 test samples each of which contains social background of a person. We predict whether this person makes over 50K a year based on its background. After one-hot coding for categorical attributes, we get 113 features of each sample as \cite{feng2018multi}. We train a GBDT-BP model with structure $113\rightarrow32\rightarrow1$.

\emph{Protein Localization}: This dataset is a 10 class classification task that contains 1484 samples. There are 8 measurements of a protein sequence in each sample. Our goal is to predict protein localization sites with 10 possible choices. We evaluate and compare the accuracies of different methods via 10-fold cross-validation. For a fair comparison, we use exactly the same set of folds as \cite{feng2018multi}. We train a GBDT-BP model using structure $8\rightarrow16\rightarrow10$. We also examine the effect of layers on this dataset. The dimension for each hidden layer is fixed to be 16. As shown in Table \ref{tab:layer}, although the test accuracies decrease as the the number of layers increases, the results are stable. The results demonstrate the effectiveness of GBDT-BP when training deeper models. 

\hl{
	\emph{Drug Consumption}: This dataset is a 7 class classification task that contains 1885 samples. There are 12 measurements of people's personality information in each sample. Our goal is to predict when drugs are used by a person: "never", "over a decade ago", "in last decade", "in last year", "in last month", "in last week", or "in last day". We evaluate and compare the accuracies of different methods via 10-fold cross-validation. We train a GBDT-BP model using structure $12\rightarrow32\rightarrow7$. 
}

\hl{
	\emph{MNIST-3000}: MNIST (Mixed National Institute of Standards and Technology) is a popular dataset for evaluating machine learning algorithms. There are 50000 training samples and 10000 test samples. Each sample is a $28 \times 28$ gray image which expresses one of digits from 0 to 9. Because training mGBDT consumes a lot of memory, we only use the first 3000 samples for training. We call this dataset \textit{MNIST-3000}. We reshape each image to a vector before fitting it into models. We train a GBDT-BP model using structure $784\rightarrow64\rightarrow10$. Experimental results are presented in Table~\ref{tab:tab}.
}

\hl{
	The proposed method achieves the best performance except on \emph{Income Prediction}. The proposed method is obviously better than single-layered GBDT. This indicates the benefits of multi-layered structure. The proposed method is obviously better than single-layered GBDT-PL. This indicates the performance gain is mainly contributed by multi-layered structure instead of the piece-wise linear function of leaves. The proposed method is competitive with or sightly better than mGBDT. This indicates the feasibility of training GBDT via back propagation.
}

\begin{figure*}[ht]
	\centering
	\subfigure[GBDT-BP]{
		\includegraphics[width=0.36\textwidth]{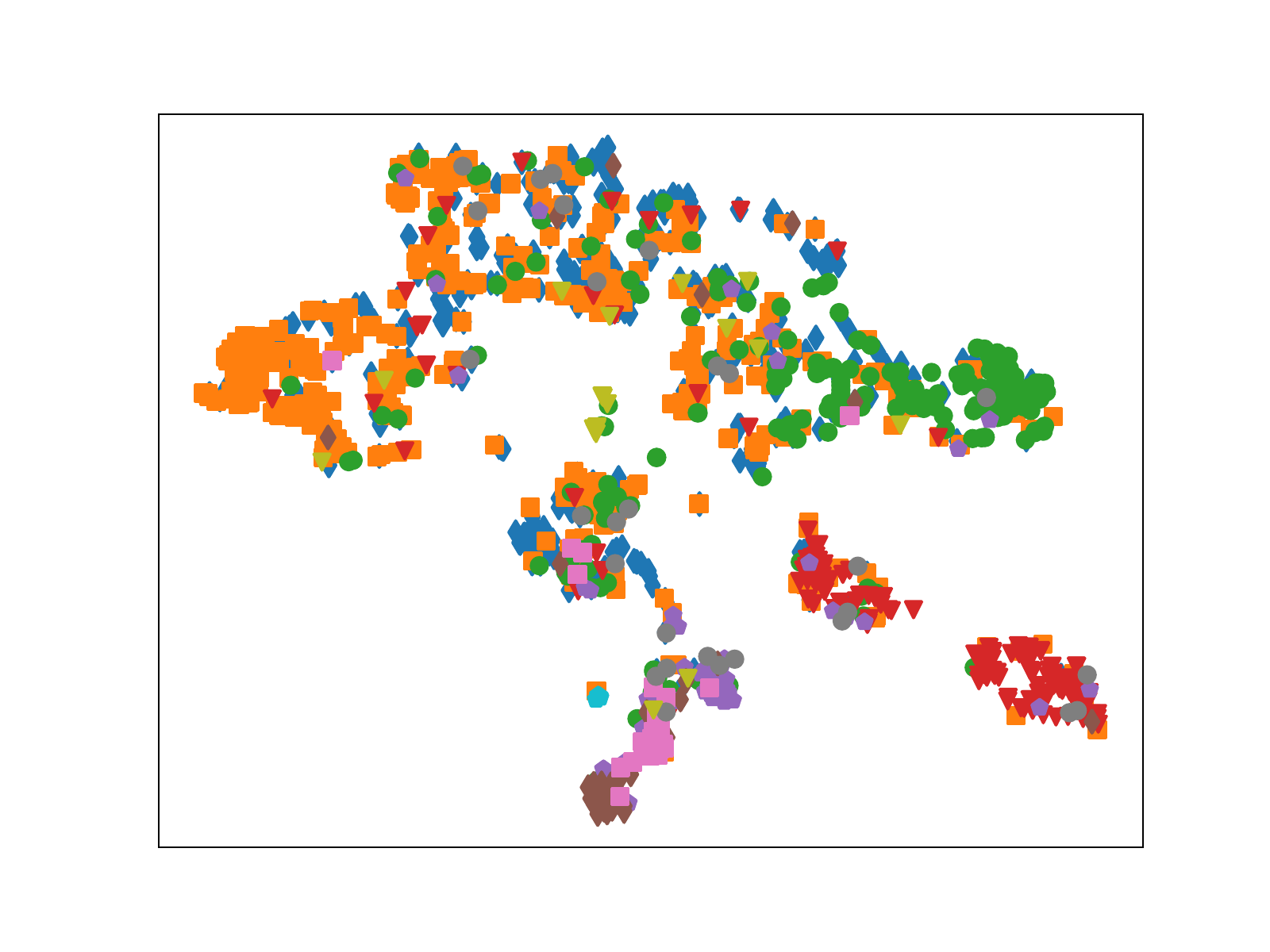}}
	\subfigure[NN]{
		\includegraphics[width=0.36\textwidth]{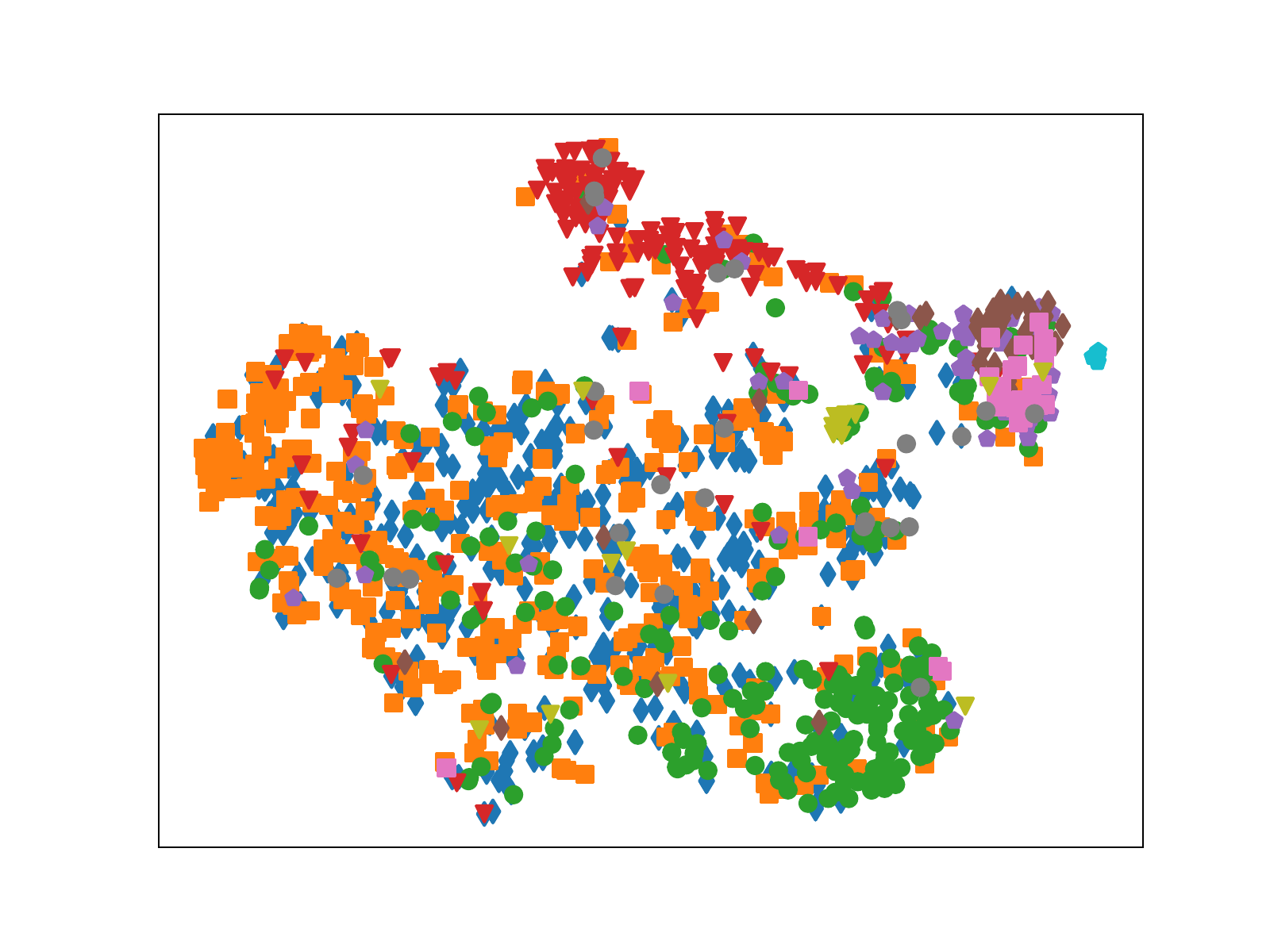}}
	\subfigure[GBDT-NN]{
		\includegraphics[width=0.36\textwidth]{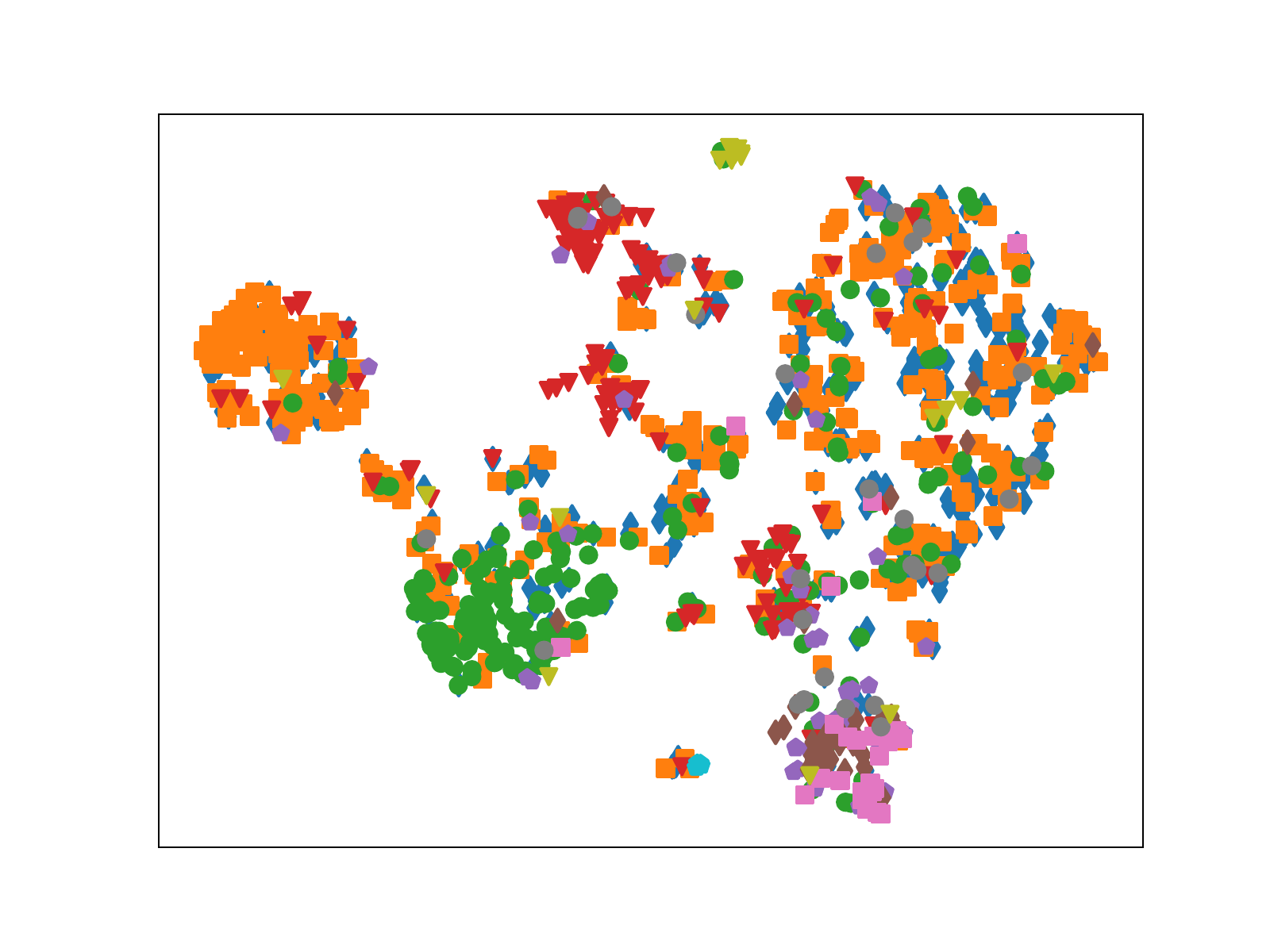}}
	\subfigure[NN-GBDT]{
		\includegraphics[width=0.36\textwidth]{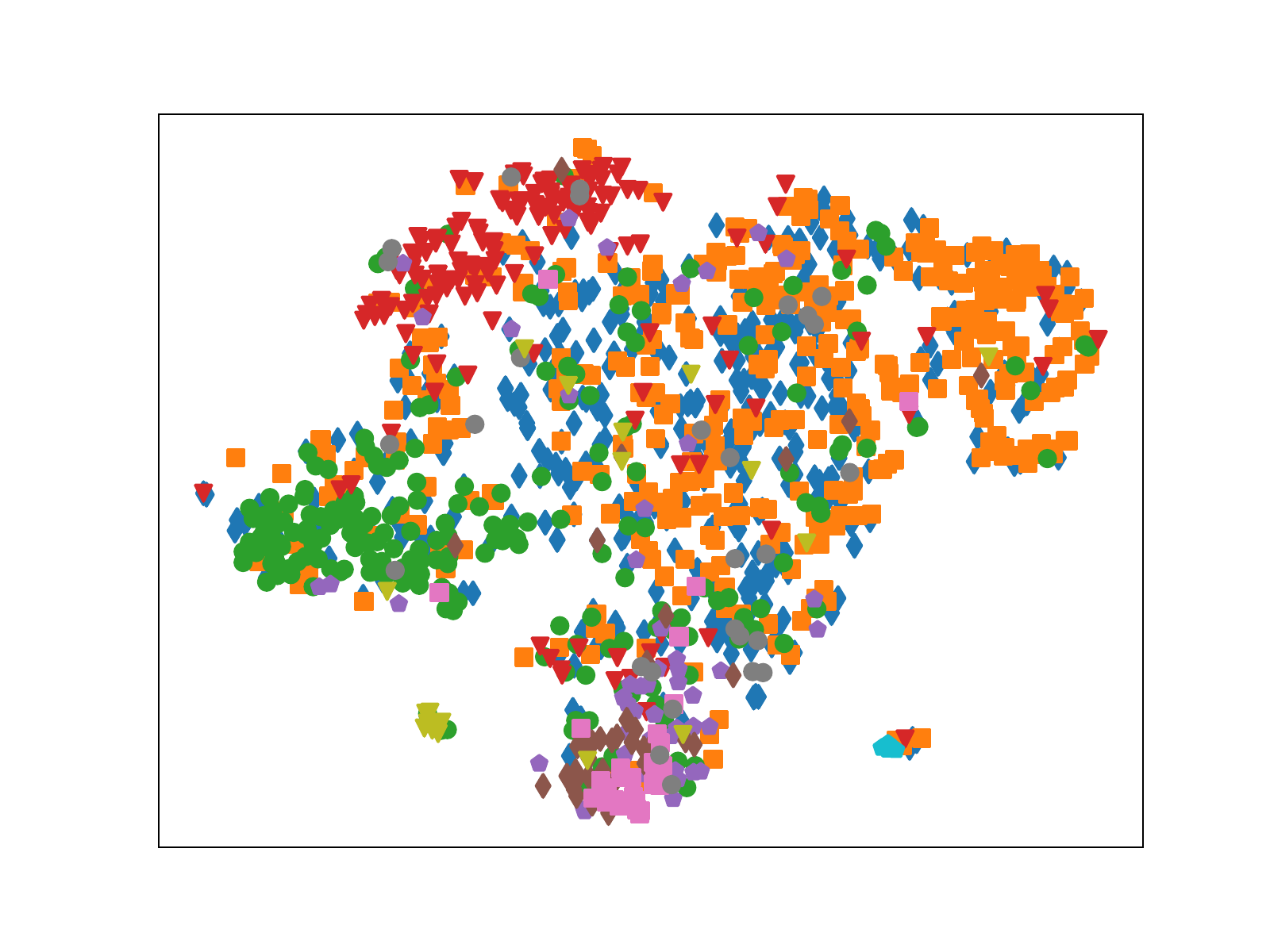}}
	\caption{\hl{We visualize the hidden representations on \emph{Protein Localization} using T-SNE. The clusters of GBDT-BP have larger margin compared with NN. The representations of GBDT-NN is similar to GBDT-BP and the representations of NN-GBDT is similar to NN.}}
	\label{fig:vis}
\end{figure*}

\begin{table}[t]
	\small
	\caption{\hl{Classification accuracy on \textit{Protein Localization}.}}
	\label{tab:mix}
	\begin{center}
		\begin{tabular}{lll}
			\toprule
			Method &  \textit{Income Prediction} & \textit{Protein Localization} \\
			\midrule
			NN                & 0.8543   & 0.5907 $\pm$ 0.0268 \\
			GBDT-BP    & 0.8741    & 0.6186 $\pm$ 0.0296 \\
			NN-GBDT   & 0.8611   & 0.6020 $\pm$ 0.0257 \\
			GBDT-NN   & \textbf{0.8748}   & \textbf{0.6203 $\pm$ 0.0281} \\
			\bottomrule
		\end{tabular}
	\end{center}
\end{table}

\subsection{Combination of GBDT and NN}
\hl{
	Back propagation with gradient descant becomes a dominant learning paradigm of neural networks. Training GBDT with back propagation makes it possible to jointly train a model which combines GBDT and NN. This provides more flexibility to design novel machine learning models.
}

\hl{
	There are two natural ways to combine GBDT and NN. We first construct GBDT layers, then we construct NN layers following GBDT layers. We call it GBDT-NN. Or we first construct NN layers, the we construct GBDT layers following NN layers. We call it NN-GBDT. In this section, we evaluate GBDT-NN and NN-GBDT on \emph{Income Prediction} and \emph{Protein Localization}. We train models with structure $113 \rightarrow 32 \rightarrow 1$ on \emph{Income Prediction} and models with structure $8 \rightarrow 32 \rightarrow 10$ on \emph{Protein Localization}. For NN-GBDT, the NN layer is updated by 15 stochastic gradient descant steps in each epoch.
}

\hl{
	Experiments are shown in Table~\ref{tab:mix}. The results indicate the feasibility of combining NN and GBDT. GBDT-NN achieves the best performance. GBDT-NN is much better than NN-GBDT. One possible reason is that the approximated gradient by GBDT-BP is not accurate and it may mislead the updating of NN. Better gradient estimation algorithms of GBDT are helpful in the future. We also visualize the hidden representations on \emph{Protein Localization} using T-SNE in Figure~\ref{fig:vis}. The clusters of GBDT-BP have larger margin compared with NN. This indicates that GBDT-BP learns better hidden representations than NN. Moreover, the representations of GBDT-NN is similar to GBDT-BP and the representations of NN-GBDT is similar to NN. This indicates that the earlier layers may have greater impact on the representations.
}

\section{Discussions}
mGBDT is the most related work to GBDT-BP. We highlight two differences between them: 1) The main difference is how hidden variables are updated. mGBDT updates them via target propagation by fitting a set of inverse functions; 2) The final layer of mGBDT is a fully connected neural network layer. The mixture of GBDT and neural network makes it difficult to recognize the contribution of each part. When we train GBDT-BP with a smaller batch size than the full batch size, the performance is slightly decreased. Thus, we train GBDT-BP with full batch size in this work. As mentioned in Section~\ref{sec:overall}, once the hidden representations are updated, we re-train GBDTs from scratch. An alternative way is that we keep previously trained GBDTs and train new GBDTs to fit the change of hidden representations. Our consideration is that the change of hidden representations is not consistent, thus leading to the redundancy of trees. Training GBDT-BP is time-consuming because 1) it requires many epochs; and 2) it requires multiple GBDTs in hidden layers. For the first reason, we may need better updating rules. For the second reason, we may use recently proposed GBDT-MO \cite{zhang2020gbdt} to learn multiple outputs simultaneously. We leave faster training as future works.

\section{Conclusions and Future Works}
We have proposed a method of learning multi-layered GBDT via back propagation. We have achieved this by replacing the constant leaf value with linear regression and approximating its derivatives. We have provided new probabilities of learning deep tree based ensembles. Experiments on both synthetic and real-world datasets verify that the proposed method outperforms single-layered GBDT in terms of both performance and representation ability. Moreover, the proposed method supports both supervised and unsupervised learning.


\newpage
{\small
	\bibliographystyle{ieee}
	\bibliography{mgbdt}
}
\end{document}